\def\prolandarxiv{1}
\documentclass{article}
\ifdefined\prolandarxiv
\usepackage[parfill]{parskip}
\usepackage{amsmath,amssymb,amsthm}
\usepackage{iftex}
\ifPDFTeX
  \usepackage[T1]{fontenc}
  \usepackage[tt=false,type1=true]{libertine}
  \usepackage[varqu]{zi4}
\else
  \usepackage{fontspec}
  \let\prolandtextfamily\rmdefault
  \renewcommand{\rmdefault}{LinuxLibertineT-TLF}
\fi
\usepackage[libertine]{newtxmath}
\ifPDFTeX\else
  \let\rmdefault\prolandtextfamily
\fi
\usepackage{bm}
\usepackage{natbib}
\setcitestyle{authoryear,round,citesep={;},aysep={,},yysep={;}}
\PassOptionsToPackage{colorlinks=true,linkcolor=blue!50!black,
  citecolor=green!40!black,urlcolor=blue!50!black}{hyperref}

\date{}

\usepackage{authblk}
\else
\usepackage{iclr2027_conference,times}
\fi
\usepackage{amsmath,amssymb,amsthm,mathtools,booktabs,array,tabularx}
\usepackage{algorithm,algpseudocode}
\usepackage[table]{xcolor}
\usepackage{multirow}
\definecolor{prolandrow}{HTML}{F0F3F6}
\newcolumntype{Y}{>{\raggedright\arraybackslash}X}
\usepackage[T1]{fontenc}
\newcommand{\method}{ProLand}
\newcommand{\R}{\mathbb R}
\newcommand{\E}{\mathbb E}
\newcommand{\Ezero}{\mathbb E_0}
\newcommand{\Fcal}{\mathcal F}
\newcommand{\Sspace}{\mathcal S}
\newcommand{\Diff}{\mathord{\mathrm D}\mkern1mu}
\newcommand{\St}{\operatorname{St}}
\newcommand{\rank}{\operatorname{rank}}
\newcommand{\range}{\operatorname{range}}
\newcommand{\polar}{\operatorname{polar}}
\newcommand{\sym}{\operatorname{sym}}
\newcommand{\grad}{\operatorname{grad}}

\newcommand{\op}{\mathrm{op}}
\newcommand{\fro}{\mathrm F}
\newcommand{\ip}[2]{\langle #1,#2\rangle_{\fro}}
\newcommand{\nrm}[1]{\lVert #1\rVert}
\newcommand{\Hbar}{\bar{H}}
\newcommand{\Amax}{A_{\max}}

\newtheorem{theorem}{Theorem}[section]
\newtheorem{proposition}[theorem]{Proposition}
\newtheorem{lemma}[theorem]{Lemma}

\theoremstyle{definition}
\newtheorem{assumption}{Assumption}

\usepackage{graphicx}

\usepackage{needspace}
\usepackage{hyperref,url}
\usepackage{placeins}
\usepackage{xcolor}

\hypersetup{pdftitle={\method{}: A New Infeasible Manifold Optimization Method for LoRA Fine-Tuning},pdfauthor={},pdfsubject={}}
\title{\method{}: A New Infeasible Manifold Optimization Method for LoRA Fine-Tuning}
\ifdefined\prolandarxiv
\author[1,*]{Chenyu Zhou}
\author[1,*]{Yuxuan Zeng}
\author[1]{Qi Deng}
\author[1]{Dongdong Ge}
\author[1,2,3]{Yinyu Ye}
\affil[1]{Shanghai Jiao Tong University}
\affil[2]{Shanghai Institute for Mathematics and Interdisciplinary Sciences (SIMIS)}
\affil[3]{Stanford University}
\affil[ ]{%
  \href{mailto:chenyuzhou@sjtu.edu.cn}{\texttt{chenyuzhou@sjtu.edu.cn}}\quad
  \href{mailto:zengyuxuan0522@sjtu.edu.cn}{\texttt{zengyuxuan0522@sjtu.edu.cn}}\quad
  \href{mailto:qdeng24@sjtu.edu.cn}{\texttt{qdeng24@sjtu.edu.cn}}\quad
  \href{mailto:ddge@sjtu.edu.cn}{\texttt{ddge@sjtu.edu.cn}}}
\affil[ ]{%
  \href{mailto:yinyu-ye@stanford.edu}{\texttt{yinyu-ye@stanford.edu}}}
\makeatletter
\g@addto@macro\@thanks{\footnotetext[1]{Equal contribution. The order of the first two authors may be interchanged.}}
\makeatother
\hypersetup{pdfauthor={Chenyu Zhou, Yuxuan Zeng, Qi Deng, Dongdong Ge, Yinyu Ye}}

\else
\author{Anonymous authors}
\fi
\begin{document}
\maketitle
\begin{abstract}
Low-rank adaptation (LoRA) enables parameter-efficient fine-tuning but can suffer from redundant basis directions. Constraining the basis factor to the Stiefel manifold addresses this redundancy through orthonormality. Landing methods allow temporarily infeasible iterates to explore a neighborhood of the manifold while progressively restoring orthogonality without per-step retractions. To improve optimization efficiency while retaining this flexibility, we introduce ProLand (Product-space Landing), a retraction-free optimizer that combines coordinated factor updates with Muon-style spectral directions. 
By designing the updates in the space of the represented weight increment, ProLand separates objective optimization from orthogonality restoration, supporting efficient feasibility recovery. This coordinated construction also enables Muon-style directions to guide objective optimization while preserving that separation. Experiments cover four GLUE validation tasks on DeBERTa-v3-base and seven language benchmarks on Qwen2.5-3B, Llama-3.2-3B, and Moonlight-16B-A3B across multiple adapter ranks. Across these 11 tasks, ProLand achieves the highest average score among the compared methods in every tested encoder and decoder model--rank configuration.
\end{abstract}
\section{Introduction}\label{sec:introduction}

Low-rank adaptation (LoRA) makes fine-tuning large language models more accessible by learning a low-rank weight increment $\Delta W=BA$ while keeping pretrained weights fixed~\citep{hu2021lora}. 
Its parameter efficiency depends on making effective use of a small number of trainable directions. The columns of $B$ define these directions in the output space, while $A$ contains their coordinates~\citep{zhu2024asymmetry}. However, \citet{park2025stiefellora} observe that AdamW fine-tuning can produce highly correlated columns of $B$, leading to basis redundancy and inefficient use of the available rank. They address this issue by imposing $B^\top B=I$, so that the basis is optimized on the Stiefel manifold. This constraint keeps the basis directions orthonormal while leaving their coordinates free to adapt. It preserves the class of representable rank-at-most-$r$ weight increments, as discussed in Section~\ref{sec:preliminaries}.

Once the Stiefel constraint is imposed, a central algorithmic choice is how to maintain orthogonality during optimization. Stiefel-LoRA~\citep{park2025stiefellora} enforces orthonormality through a retraction at each step. Landing methods~\citep{ablin2022landing,ablin2024stiefel} instead allow temporarily infeasible iterates and use an explicit correction to restore feasibility, avoiding per-step retractions. Manifold-LoRA~\citep{zhang2026manifoldlora} applies this approach to LoRA fine-tuning. Beyond avoiding repeated retractions, allowing controlled departures from the manifold gives the optimizer additional freedom to explore the surrounding space. Figure~\ref{fig:motivation} illustrates the potential benefit by comparing our proposed method (ProLand) with variants that replace its Landing correction with per-step retractions to enforce $B^\top B=I$. From the same initialization and under the same rule for computing product-space directions at their respective iterates, ProLand reaches a better stationary solution in this example. This observation motivates exploring infeasible paths while retaining control over constraint violation.

Our goal is to improve the efficiency of this form of infeasible optimization for LoRA: make effective progress on the fine-tuning objective while rapidly recovering feasibility. These two goals are coupled, since an update that reduces the loss can also perturb the orthogonality of $B$ and interfere with the Landing correction. At the same time, the model depends on the joint weight increment $BA$; hence the directions chosen for the two factors should be coordinated. We introduce \textbf{ProLand} (Product-space Landing), a retraction-free optimizer that addresses these issues through two complementary design choices.

\textbf{First, coordinated updates in the product space support rapid feasibility recovery.} We design the joint first-order change $D_BA+BD_A$ in the represented weight increment and realize it through factor directions satisfying $B^\top D_B=0$, even when $B$ is infeasible. This condition makes the optimization step leave $B^\top B$ unchanged to first order and eliminates its first-order interaction with the Landing correction. Consequently, the optimization step perturbs the constraint violation only at second order in its step size. Under the conditions of Section~\ref{sec:theory}, this violation contracts geometrically toward a neighborhood whose radius is quadratic in that step size. This provides a mechanism for recovering feasibility while continuing to optimize the represented weight increment.

\textbf{Second, Muon-style directions guide objective optimization.} We incorporate the spectral update principle of Muon~\citep{jordan2024muon} into this coordinated construction. ProLand forms a momentum input in the space of realizable first-order changes in $BA$, computes its polar factor, and projects the result back into that space. 
The spectral direction is therefore defined at the level of \(BA\), rather than independently in the two factor spaces, and is realized by the factor updates above while retaining their separation from feasibility restoration.
This allows the spectral update and the Landing correction to be combined without a retraction at every step.

For the resulting method, we establish joint guarantees for objective stationarity and feasibility under suitable regularity and step-size conditions, assuming exact arithmetic. We show that the deterministic ProLand method attains $O(T^{-1/2})$ product-space stationarity and $O(T^{-1})$ constraint violation at a randomized output. With stochastic gradients and horizon-dependent momentum and step size, the corresponding expected bounds are $O(T^{-1/4})$ and $O(T^{-3/2})$. The constraint-violation bounds also hold uniformly along each trajectory after an initial transient.

To make this construction practical for LoRA fine-tuning, we derive an exact low-rank implementation using momentum factors of width at most $2r$ and polar-factor cores of size at most $2r\times2r$, avoiding dense weight-sized optimizer states. In experiments, ProLand achieves the highest task-averaged scores among the compared methods on four GLUE tasks with DeBERTa-v3-base and across all tested decoder model--rank configurations. Matched ablations show that the Landing correction substantially reduces constraint violation.

\subsection{Related Work and Positioning}\label{sec:related}

Several methods improve LoRA optimization by coordinating the updates of its two factors. LoRA-Pro~\citep{wang2025lorapro} and LORO~\citep{mo2025loro} account for the change in the represented matrix when designing factor updates for fine-tuning and low-rank pretraining, respectively. LoRA-Muon~\citep{cesista2026loramuon} and sMuon~\citep{anson2026smuon} extend Muon-style spectral optimization to low-rank adapters. In particular, sMuon reconstructs a low-rank gradient from factor gradients and derives coordinated directions through linearization and least squares, with momentum represented in factor coordinates. ProLand additionally imposes the Stiefel constraint on $B$ and selects a factor realization satisfying $B^\top D_B=0$ at both feasible and infeasible iterates. This realization enables its spectral update to work with a separate feasibility correction.

Spectral updates have also been developed explicitly for manifold optimization. Riemannion~\citep{bogachev2026riemannion} extends Muon to optimization of the represented matrix on the fixed-rank manifold. LoRA-TSD~\citep{andriianov2026loratsd} develops tangent-space spectral descent on that manifold and maps its updates back to the factors through a factor-induced retraction. Skewon~\citep{solonko2026skewon} constructs spectral directions for a Stiefel-constrained matrix by solving a linear minimization problem over tangent directions subject to a spectral-norm constraint, followed by a retraction. These methods illustrate how spectral updates can be adapted to manifold geometry. ProLand focuses on the complementary challenge of combining such directions with infeasible iterates and controlled restoration of the basis constraint.

A complementary line of work avoids per-step retractions through Landing methods, which allow infeasible iterates and restore orthogonality through an explicit correction~\citep{ablin2022landing,ablin2024stiefel}. Manifold-LoRA~\citep{zhang2026manifoldlora} applies this approach to Stiefel and oblique constraints. Its update construction uses the gradient with respect to $A$ and a projected-gradient expression for $B$, together with a feasibility correction; its analysis establishes convergence of the underlying retraction-free gradient method. ProLand retains the Landing mechanism and constructs its optimization direction in the space of joint changes in $BA$. The resulting factor directions satisfy $B^\top D_B=0$ even away from the manifold, eliminating all terms linear in the optimization step size from the recurrence for $B^\top B-I$. For this construction, we establish product-space stationarity guarantees and pathwise bounds on constraint violation under the conditions in Section~\ref{sec:theory}. Appendix~\ref{app:prior} gives further comparisons.
\section{Problem Setup and Preliminaries}\label{sec:preliminaries}

% \paragraph{LoRA formulation.}
Consider a single LoRA module with all other model parameters held fixed.
We parameterize its trainable weight increment as $\Delta W=BA$, where $B\in\R^{m\times r}$, $A\in\R^{r\times n}$, and $1\le r\le\min\{m,n\}$.
Let $f(\Delta W)$ denote the corresponding fine-tuning objective.
Imposing the orthonormality constraint $B^\top B=I_r$ on $B$ while leaving $A$ unconstrained~\citep{park2025stiefellora,zhang2026manifoldlora} yields
\begin{equation}\label{eq:problem}
\min_{B\in\St(m,r),\,A\in\R^{r\times n}} f(BA),
\end{equation}
where $\St(m,r):=\{B\in\R^{m\times r}:B^\top B=I_r\}$ denotes the Stiefel manifold.

The orthonormality constraint preserves the set of representable weight increments:
\[
\left\{BA:B\in\St(m,r),\ A\in\R^{r\times n}\right\}
=
\left\{\Delta W\in\R^{m\times n}:\rank(\Delta W)\le r\right\}.
\]
Thus, the constraint restricts the factorization, not the class of rank-at-most-$r$ weight increments. A proof is given in Appendix~\ref{app:product-geometry}.

At a feasible point $B\in\St(m,r)$, the tangent space $T_B\St(m,r)$ consists of matrices $D\in\R^{m\times r}$ satisfying $B^\top D+D^\top B=0$~\citep{absil2008matrix}. Accordingly, for arbitrary $B$, let $C(B):=B^\top B-I_r$ and measure the constraint violation, also called the feasibility error, by $\nrm{C(B)}_{\op}$, which is zero at feasible points.

\paragraph{First-order product geometry.}
The differential of the product map $\Phi(B,A):=BA$ is
\[
\Diff\Phi(B,A)[\Delta B,\Delta A]
=
\Delta B\,A+B\,\Delta A.
\]
The image of this differential, $\Sspace_{B,A}:=\range\Diff\Phi(B,A)$, is the corresponding first-order product space. Let $(\cdot)^\dagger$ denote the Moore--Penrose pseudoinverse, and define $P_B:=BB^\dagger$ and $P_A:=A^\dagger A$, the orthogonal projectors onto the column space of $B$ and the row space of $A$, respectively. Under the Frobenius inner product $\ip{X}{Y}:=\operatorname{tr}(X^\top Y)$, with induced norm $\nrm{\cdot}_{\fro}$, the orthogonal projector onto $\Sspace_{B,A}$ is
\begin{equation}\label{eq:projector}
\Pi_{B,A}Z=P_BZ+(I-P_B)ZP_A.
\end{equation}
The same orthogonal projection operator applies to arbitrary-rank factors, including $B\notin\St(m,r)$; see Appendix~\ref{app:product-geometry}.

For $F:=\nabla f(BA)$, define the projected gradient $R:=\Pi_{B,A}F$. Since $\Pi_{B,A}$ is an orthogonal projector, $R=0$ if and only if $F\perp\Sspace_{B,A}$. This is equivalent to $B^\top F=0$ and $FA^\top=0$; see the derivation in Appendix~\ref{app:stationarity}. When $B\in\St(m,r)$, these conditions characterize first-order stationarity of the factorized constrained problem. We use the nuclear norm $\nrm{R}_*$ as the product-space stationarity measure and quantify constraint violation separately by $\nrm{C(B)}_{\op}$.

\paragraph{Polar factor and spectral--nuclear duality.}
For a nonzero matrix $H$, let $H=U\Sigma V^\top$ be its compact singular value decomposition, retaining only nonzero singular values. We define its canonical polar factor as $\polar(H):=UV^\top$, with $\polar(0):=0$~\citep{jordan2024muon}. This choice satisfies $\nrm{\polar(H)}_{\op}\le1$.

The nuclear norm is dual to the operator norm, and the polar factor attains the corresponding dual maximum:
\[
\max_{\nrm{D}_{\op}\le1}\ip{H}{D}
=
\ip{H}{\polar(H)}
=
\nrm{H}_*.
\]
ProLand constructs its product-space update direction from this polar factor.
\section{The \method{} Method}\label{sec:method}

ProLand combines a product-space spectral update with a Landing correction. Its construction addresses three geometric issues and one implementation issue. First, as the factors change, past momentum need not lie in the current product space. Second, even an input in this space may have a polar factor outside it. Third, the factor directions must realize the selected first-order product direction while allowing a separate correction to control constraint violation. Finally, the product-space operations should admit an exact low-rank implementation.

At iteration $t$, let $G_t$ be a stochastic estimate of $\nabla f(B_tA_t)$. Algorithm~\ref{alg:proland} gives the complete iteration. Both updates are computed at $(B_t,A_t)$ and applied simultaneously.

\begin{algorithm}[H]
\caption{\method{}}
\label{alg:proland}
\begin{algorithmic}[1]
\Require Initial factors $B_0,A_0$; initial momentum $\Hbar_{-1}$; iteration budget $T$; momentum coefficient $\beta\in[0,1)$; optimization step sizes $\{\gamma_t\}_{t=0}^{T-1}$; Landing step size $\rho>0$
\For{$t=0,\ldots,T-1$}
\State Obtain $G_t$ at $B_tA_t$; set $\Pi_t\gets\Pi_{B_t,A_t}$.
\State $\Hbar_t\gets\beta\Pi_t\Hbar_{t-1}+(1-\beta)\Pi_tG_t$.
\State $J_t\gets\beta\Hbar_t+(1-\beta)\Pi_tG_t$; $Y_t\gets\polar(J_t)$.
\State $D_{A,t}\gets B_t^\dagger Y_t$.
\State $D_{B,t}\gets(I-P_{B_t})Y_tA_t^\dagger$.
\State $C_t\gets B_t^\top B_t-I_r$.
\State $A_{t+1}\gets A_t-\gamma_tD_{A,t}$;
\Statex \hspace{\algorithmicindent}$B_{t+1}\gets B_t-\gamma_tD_{B,t}-\rho B_tC_t$. \hfill $\triangleright$ simultaneous update
\EndFor
\end{algorithmic}
\end{algorithm}

\paragraph{Reprojected momentum.}
Write $\Sspace_t:=\Sspace_{B_t,A_t}$ and $R_t:=\Pi_t\nabla f(B_tA_t)$. ProLand forms $\Hbar_t$ from the reprojected momentum $\Pi_t\Hbar_{t-1}$ and the projected gradient $\Pi_tG_t$, a stochastic estimate of $R_t$. Both quantities lie in $\Sspace_t$; hence $\Hbar_t\in\Sspace_t$. Similarly, $J_t\in\Sspace_t$ because it combines $\Hbar_t$ with the same projected gradient.

\paragraph{Product-space polar direction.}
Although $J_t\in\Sspace_t$, its polar factor $Y_t$ need not lie in this space. ProLand therefore selects the first-order product direction $\xi_t:=\Pi_tY_t$.

\paragraph{Factor realization and Landing.}
The factor directions in Algorithm~\ref{alg:proland} realize this projection without explicitly forming $\xi_t$:
\begin{equation}\label{eq:stiefel-compatible-realization}
B_tD_{A,t}+D_{B,t}A_t=\xi_t,
\qquad
B_t^\top D_{B,t}=0.
\end{equation}
The orthogonality condition holds even when $B_t$ is infeasible and implies $D_{B,t}\in T_{B_t}\St(m,r)$ at feasible points. This condition is not restrictive: every direction in $\Sspace_t$ admits a factor realization satisfying it; see Appendix~\ref{app:factor-realization}.

To control constraint violation without per-step retractions, ProLand adds the Landing correction $-\rho B_tC_t$ to the $B$-update. This is a negative gradient step for $\mathcal N(B):=\tfrac14\nrm{C(B)}_{\fro}^2$, since $\nabla\mathcal N(B_t)=B_tC_t$. This correction is also compatible with the optimization direction: the condition $B_t^\top D_{B,t}=0$ gives $\ip{D_{B,t}}{B_tC_t}=0$; that is, the optimization direction and the Landing correction are Frobenius-orthogonal, and their step sizes $\gamma_t$ and $\rho$ are controlled separately.

\begin{figure}[t]
\centering
\includegraphics[width=\linewidth]{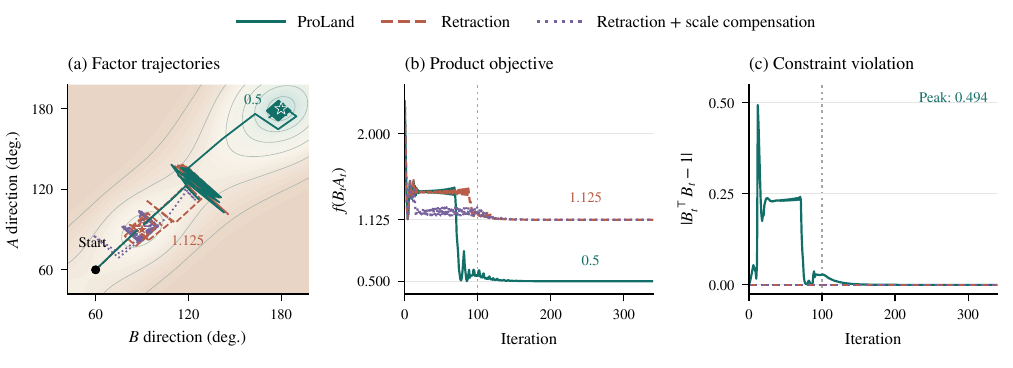}
\caption{
\textbf{Feasibility treatment changes the solution.}
(a) Factor trajectories over loss contours minimized over the product scale;
(b) Product objective values;
(c) Constraint violation.
}
\label{fig:motivation}
\end{figure}

In Figure~\ref{fig:motivation}, replacing this Landing update with per-step retractions from the same initialization and under the same product-space direction rule leads to distinct stationary solutions, while all variants attain feasibility to numerical tolerance. Further comparison details are given in Appendix~\ref{app:motivation}.

\paragraph{Exact low-rank implementation.}
ProLand admits an exact low-rank implementation. Since every matrix in $\Sspace_t$ has rank at most $2r$, both $\Hbar_t$ and $J_t$ admit thin factorizations. Given an initial momentum factorization of width at most $2r$, the recursion operates on these factors and evaluates the polar factor through a core of dimensions at most $2r\times2r$. The factor gradients $g_{A,t}=B_t^\top G_t$ and $g_{B,t}=G_tA_t^\top$ from the same sampled loss suffice for the recursion, without forming $G_t$. Excluding gradient evaluation, each iteration requires $O((m+n)r^2+r^3)$ arithmetic operations and $O((m+n)r)$ additional storage. Appendix~\ref{app:backend} gives the implementation and proves its equivalence to Algorithm~\ref{alg:proland} in exact arithmetic.

\FloatBarrier
\section{Theoretical Analysis}\label{sec:theory}
\begingroup
\renewcommand{\theassumption}{\thesection.\arabic{assumption}}

Section~\ref{sec:method} shows that ProLand realizes the target first-order product direction through coordinated factor updates. We analyze Algorithm~\ref{alg:proland} in exact arithmetic, first establishing feasibility and factor conditioning and then deriving product-space stationarity guarantees.

\begin{assumption}[Objective and stochastic regularity]\label{ass:objective}
The objective $f$ is bounded below and $L$-smooth: $f(W)\ge f_*>-\infty$ and $\nrm{\nabla f(W)-\nabla f(W')}_{\fro}\le L\nrm{W-W'}_{\fro}$ for all $W,W'$. In the stochastic setting, let $\{\Fcal_t\}$ be the filtration generated by the algorithmic history up to iteration $t$, before sampling $G_t$. For the stochastic gradient $G_t$ defined in Section~\ref{sec:method}, assume $\E[G_t\mid\Fcal_t]=\nabla f(B_tA_t)$ and $\E[\nrm{G_t-\nabla f(B_tA_t)}_{\fro}^2\mid\Fcal_t]\le\sigma^2$ for a finite constant $\sigma\ge0$. These are standard assumptions in stochastic nonconvex optimization~\citep{ghadimi2013stochastic,bottou2018optimization}.
\end{assumption}

Since LoRA commonly uses a zero-initialized factor, we impose the following regularity condition only after a finite initialization phase and relabel the first iterate of this regular segment as $t=0$~\citep{hayou2024initialization,wang2025lorapro}.

\begin{assumption}[$A$-factor regularity]\label{ass:factor}
There exist constants $a_->0$ and $\Amax<\infty$, independent of $T$, such that $a_-\le\sigma_r(A_t)\le\nrm{A_t}_{\op}\le\Amax$ for $0\le t\le T$. In the stochastic setting, these bounds hold almost surely with common deterministic constants.
\end{assumption}

Similar full-rank and uniform conditioning assumptions are used in analyses of low-rank factorization~\citep{wang2025lorapro,savostianova2023robust}. Here, the uniform lower bound keeps $A_t^\dagger$ bounded, enabling uniform estimates for the factor realization and the variation of the product-space projector $\Pi_t$; see Appendices~\ref{app:Awindow} and~\ref{app:tracking}.

We first establish pathwise feasibility and the resulting conditioning of $B_t$.

\begin{proposition}[Feasibility and $B$-factor conditioning]\label{thm:feasibility}
Let Algorithm~\ref{alg:proland} satisfy Assumption~\ref{ass:factor}. Fix $0<\epsilon<1$ and $0<\rho\le1/2$, suppose $\nrm{C_0}_{\op}\le\epsilon$, and define $q_C:=1-(1-\epsilon)(2\rho+\rho^2\epsilon)\in(0,1)$.
If the positive step sizes satisfy $\sup_{0\le t<T}\gamma_t\le a_-\sqrt{\epsilon(1-q_C)}$, then, for all $0\le t\le T$,
\[
\nrm{C_t}_{\op}\le\epsilon,\qquad
\sqrt{1-\epsilon}\le\sigma_r(B_t)\le\sigma_1(B_t)\le\sqrt{1+\epsilon},\qquad
\nrm{B_t^\dagger}_{\op}\le(1-\epsilon)^{-1/2}.
\]
Moreover, for constant $\gamma_t\equiv\gamma$,
\begin{equation}\label{eq:feasbound}
\nrm{C_t}_{\op}
\le
q_C^t\nrm{C_0}_{\op}
+
\frac{\gamma^2(1-q_C^t)}{a_-^2(1-q_C)}.
\end{equation}
\end{proposition}

The mechanism behind Proposition~\ref{thm:feasibility} follows from the exact Gram recursion $C_{t+1}=(1-2\rho)C_t+(\rho^2-2\rho)C_t^2+\rho^2C_t^3+\gamma_t^2D_{B,t}^{\top}D_{B,t}$. The identity $B_t^\top D_{B,t}=0$ cancels the terms linear in $\gamma_t$, including those coupling the optimization step and the Landing correction. Within the safe region, the Landing correction contracts the constraint violation, while the optimization step contributes only a quadratic perturbation, yielding the geometric transient and the $O(\gamma^2)$ term in the bound above. Detailed results, proofs, and finite-precision diagnostics are given in Appendices~\ref{app:feasibility} and~\ref{app:horizontal-check}.

Together with Assumption~\ref{ass:factor}, these bounds provide the uniform factor conditioning needed to control the factor realization and the variation of $\Pi_t$.

We next turn to product-space stationarity.
Spectral--nuclear duality and the self-adjointness of $\Pi_t$ give $\langle\nabla f(B_tA_t),\xi_t\rangle_{\fro}=\langle R_t,Y_t\rangle_{\fro}\ge\nrm{R_t}_*-2\nrm{J_t-R_t}_*$. This inequality supplies the leading descent term, with an error determined by the polar-factor input mismatch $J_t-R_t$. Combining it with smoothness and bounds on projector variation and momentum tracking yields the following convergence rates.

\begin{theorem}[Deterministic and stochastic convergence of ProLand]\label{thm:convergence}
Suppose Assumptions~\ref{ass:objective}--\ref{ass:factor} and the conditions of Proposition~\ref{thm:feasibility} hold, with $\epsilon,\rho$, all regularity constants, and all initial-state bounds independent of $T$. Define $\tau_T(b):=\lceil2b\log T/|\log q_C|\rceil$ for $b>0$. The following statements hold for all sufficiently large $T$.

\textbf{(i) Deterministic case.}\label{thm:deterministic} Use exact gradients, a fixed, horizon-independent momentum coefficient $\beta\in[0,1)$, and $\gamma_t\equiv\gamma_T=\theta T^{-1/2}$ for fixed $\theta>0$. Assume $\nrm{\Hbar_0-R_0}_{\fro}^2\le U_0<\infty$ for a deterministic constant $U_0$ independent of $T$. Draw $\widehat t\sim\operatorname{Unif}\{\tau_T(1/2),\ldots,T-1\}$ and return $(B_{\widehat t},A_{\widehat t})$. Then
\begin{equation}\label{eq:det-rates}
\E_{\widehat t}\nrm{R_{\widehat t}}_*
=O(T^{-1/2}),
\qquad
\E_{\widehat t}\nrm{C_{\widehat t}}_{\op}
=O(T^{-1}).
\end{equation}

\textbf{(ii) Stochastic case.}\label{thm:stochastic} Assume $\E[\nrm{\Hbar_0-R_0}_{\fro}^2\mid\Fcal_0]\le U_0<\infty$ almost surely, for a deterministic constant $U_0$ independent of $T$. Set $1-\beta_T=c_\beta T^{-1/2}$ and $\gamma_t\equiv\gamma_T=c_\gamma T^{-3/4}$ for fixed $c_\beta,c_\gamma>0$, keeping both parameters constant during the run. Draw $\widehat t\sim\operatorname{Unif}\{\tau_T(3/4),\ldots,T-1\}$, independently of the algorithmic randomness, and return $(B_{\widehat t},A_{\widehat t})$. Then
\begin{equation}\label{eq:stoch-rates}
\E\nrm{R_{\widehat t}}_*
=O(T^{-1/4}),
\qquad
\E\nrm{C_{\widehat t}}_{\op}
=O(T^{-3/2}).
\end{equation}
\end{theorem}

Thus, both guarantees apply to the same randomized output. In (i), expectation is over the output index; in (ii), it also includes the stochastic gradients. All constants hidden in $O(\cdot)$ are independent of $T$. Appendix~\ref{app:convergence} provides the full-horizon averaged-stationarity bounds and uniform pathwise constraint-violation bounds on the tail, together with the corresponding proofs.
\endgroup
\section{Experiments}\label{sec:experiments}
\subsection{Experimental Setup}\label{sec:experimental-setup}
We begin with the encoder DeBERTa-v3-base~\citep{he2023debertav3}, evaluated at rank 16 on four GLUE~\citep{wang2018glue} validation tasks: RTE, MRPC, STS-B, and CoLA. Across both encoder and decoder evaluations, we compare \method{} with AdamW, LoRA-Muon, fixed-step Manifold-LoRA, per-factor Muon, Riemannion, and sMuon. For DeBERTa, we report means and sample standard deviations across three seeds.

We then evaluate Qwen2.5-3B~\citep{qwen2024qwen25} and Llama-3.2-3B~\citep{grattafiori2024llama3,meta2024llama32} at adapter ranks $r\in\{4,8,16\}$, and Moonlight-16B-A3B~\citep{liu2025muonscalable} at rank 16, on seven language benchmarks: MMLU~\citep{hendrycks2021mmlu}, PIQA~\citep{bisk2020piqa}, WinoGrande~\citep{sakaguchi2020winogrande}, OpenBookQA~\citep{mihaylov2018openbookqa}, HellaSwag~\citep{zellers2019hellaswag}, BoolQ~\citep{clark2019boolq}, and ARC-Challenge~\citep{clark2018arc}.
The model set spans the AdamW-pretrained Qwen and Llama backbones and the Muon-pretrained Moonlight. Each task uses a separate adapter on a frozen backbone. The decoder tables report zero-shot accuracy and its unweighted mean across the seven tasks.
All experimental settings are collected in Appendix~\ref{app:experiments}.

\subsection{Encoder Results on Four GLUE Tasks}\label{sec:glue-results}
Table~\ref{tab:deberta-glue} reports results for the encoder backbone and task-specific validation metrics. \method{} achieves the highest four-task mean, $84.46\pm0.38$, exceeding the strongest baseline, sMuon, by 0.57 points, and attains the highest means on RTE and CoLA. The construction therefore transfers beyond decoder language models.

\begin{table}[H]
\centering
\caption{\textbf{DeBERTa-v3-base on GLUE validation sets.} Task columns report the mean $\pm$ sample standard deviation across seeds; Avg. is the unweighted mean of the four task scores. Bold and underline mark the best and second-best means in each column.}
\label{tab:deberta-glue}
\begingroup
\renewcommand{\baselinestretch}{1}\fontsize{9}{10.5}\selectfont
\setlength{\tabcolsep}{3pt}
\renewcommand{\arraystretch}{1.16}
\begin{tabularx}{\linewidth}{@{}>{\raggedright\arraybackslash}Xrrrrr@{}}
\toprule
Method & RTE & MRPC & STS-B & CoLA & Avg. \\
\midrule
AdamW & $79.78\pm2.25$ & $90.21\pm0.47$ & $90.89\pm0.09$ & $64.80\pm1.13$ & 81.42 \\
LoRA-Muon & $83.63\pm1.82$ & $91.71\pm0.88$ & $\underline{91.74\pm0.25}$ & $67.77\pm0.82$ & 83.71 \\
Per-factor Muon & $83.27\pm2.35$ & $\mathbf{92.67\pm0.31}$ & $\mathbf{91.75\pm0.05}$ & $67.67\pm1.89$ & 83.84 \\
sMuon & $\underline{85.08\pm0.91}$ & $91.07\pm0.66$ & $91.59\pm0.09$ & $\underline{67.83\pm0.34}$ & \underline{83.89} \\
Riemannion & $83.75\pm1.44$ & $91.33\pm0.41$ & $91.68\pm0.14$ & $66.97\pm1.37$ & 83.43 \\
Manifold-LoRA & $78.58\pm1.99$ & $89.52\pm0.91$ & $90.70\pm0.12$ & $64.08\pm1.97$ & 80.72 \\
\rowcolor{prolandrow}
\textbf{\method{} (ours)} & $\mathbf{85.20\pm1.08}$ & $\underline{92.15\pm1.16}$ & $91.64\pm0.06$ & $\mathbf{68.85\pm1.02}$ & $\mathbf{84.46}$ \\
\bottomrule
\end{tabularx}
\endgroup
\end{table}

\subsection{Decoder Language Benchmark Results}
\method{} has the highest mean score in all seven reported model--rank settings (Tables~\ref{tab:qwen25-3b-language}, \ref{tab:llama32-3b-language}, and~\ref{tab:moonlight-language}), with margins of 0.46 to 1.10 percentage points over the strongest baseline in each setting. Across the seven settings, \method{} is first or second on MMLU in every setting (first in six), and in the top two on WinoGrande and ARC-Challenge in six settings each.

\begin{table}[H]
\centering
\caption{\textbf{Qwen2.5-3B: language benchmarks.} Zero-shot accuracy (\%) at ranks 4, 8, and 16. Avg. is the unweighted mean; bold and underline mark the best and second-best scores within each rank.}
\label{tab:qwen25-3b-language}
\begingroup
\renewcommand{\baselinestretch}{1}\fontsize{9}{10.5}\selectfont
\setlength{\tabcolsep}{5.5pt}
\renewcommand{\arraystretch}{1.12}
\begin{tabularx}{\linewidth}{c>{\raggedright\arraybackslash}Xrrrrrrrr}
\toprule
Rank & Method & MMLU & PIQA & Wino & OBQA & Hella & BoolQ & ARC-C & Avg. \\
\midrule
\multirow[c]{7}{*}{4} & AdamW & 65.18 & 78.73 & 69.69 & 42.80 & 74.01 & 86.15 & \underline{49.66} & 66.60 \\
 & LoRA-Muon & 63.64 & \textbf{79.38} & \underline{71.03} & 42.20 & 72.95 & \textbf{87.16} & 49.32 & 66.53 \\
 & Per-factor Muon & 64.76 & 78.67 & 70.40 & \textbf{43.80} & 74.37 & 86.67 & 48.46 & \underline{66.73} \\
 & sMuon & \underline{65.30} & 78.67 & 69.93 & 41.40 & \textbf{74.63} & 86.45 & 49.06 & 66.49 \\
 & Riemannion & 64.25 & 79.27 & 70.40 & 42.40 & 74.43 & 86.91 & 49.40 & 66.72 \\
 & Manifold-LoRA & 65.23 & \underline{79.33} & 69.06 & 42.60 & 74.33 & 86.42 & 48.55 & 66.50 \\
\rowcolor{prolandrow}
 & \textbf{\method{} (ours)} & \textbf{65.77} & 79.22 & \textbf{71.35} & \underline{43.20} & \underline{74.46} & \underline{87.06} & \textbf{51.88} & \textbf{67.56} \\
\midrule
\multirow[c]{7}{*}{8} & AdamW & 65.28 & 79.11 & 69.38 & 42.60 & 74.39 & \underline{86.61} & 49.74 & 66.73 \\
 & LoRA-Muon & 64.96 & 79.16 & 69.46 & \textbf{43.40} & 73.75 & 85.84 & 49.32 & 66.55 \\
 & Per-factor Muon & 64.36 & 78.51 & 70.09 & 42.20 & \textbf{74.65} & 86.57 & 48.29 & 66.38 \\
 & sMuon & 65.45 & 79.00 & 70.09 & 41.80 & 73.75 & 84.74 & 47.35 & 66.03 \\
 & Riemannion & 65.33 & 78.84 & \underline{70.64} & \underline{43.00} & 74.30 & 86.33 & 49.06 & 66.78 \\
 & Manifold-LoRA & \textbf{65.90} & \textbf{79.60} & 69.53 & 41.60 & 74.43 & 86.39 & \underline{51.02} & \underline{66.92} \\
\rowcolor{prolandrow}
 & \textbf{\method{} (ours)} & \underline{65.82} & \underline{79.54} & \textbf{71.43} & \textbf{43.40} & \underline{74.46} & \textbf{88.17} & \textbf{53.33} & \textbf{68.02} \\
\midrule
\multirow[c]{7}{*}{16} & AdamW & 65.15 & 78.73 & \underline{70.64} & 42.00 & \textbf{75.00} & 87.37 & 49.91 & \underline{66.97} \\
 & LoRA-Muon & 63.27 & 76.01 & 69.53 & \textbf{44.00} & 73.15 & 87.09 & 49.06 & 66.02 \\
 & Per-factor Muon & \underline{65.33} & 78.45 & 69.14 & 43.00 & 74.70 & 86.79 & 49.91 & 66.76 \\
 & sMuon & 65.31 & 79.00 & 68.98 & 42.80 & 74.68 & 87.00 & 48.81 & 66.65 \\
 & Riemannion & 64.93 & 78.94 & \underline{70.64} & 41.60 & \underline{74.88} & \textbf{87.89} & 49.83 & 66.96 \\
 & Manifold-LoRA & 61.63 & \underline{79.33} & 69.38 & \underline{43.40} & 72.98 & 86.79 & \underline{50.60} & 66.30 \\
\rowcolor{prolandrow}
 & \textbf{\method{} (ours)} & \textbf{65.95} & \textbf{79.49} & \textbf{71.19} & \underline{43.40} & 74.68 & \underline{87.83} & \textbf{52.73} & \textbf{67.89} \\
\bottomrule
\end{tabularx}
\endgroup
\end{table}
\begin{table}[H]
\centering
\caption{\textbf{Llama-3.2-3B: language benchmarks.} Zero-shot accuracy (\%) at ranks 4, 8, and 16. Avg. is the unweighted mean; bold and underline mark the best and second-best scores within each rank.}
\label{tab:llama32-3b-language}
\begingroup
\renewcommand{\baselinestretch}{1}\fontsize{9}{10.5}\selectfont
\setlength{\tabcolsep}{5.5pt}
\renewcommand{\arraystretch}{1.12}
\begin{tabularx}{\linewidth}{c>{\raggedright\arraybackslash}Xrrrrrrrr}
\toprule
Rank & Method & MMLU & PIQA & Wino & OBQA & Hella & BoolQ & ARC-C & Avg. \\
\midrule
\multirow[c]{7}{*}{4} & AdamW & 54.70 & 77.80 & 70.09 & \underline{42.80} & 73.43 & 85.17 & 47.87 & 64.55 \\
 & LoRA-Muon & 52.76 & 76.01 & \underline{72.06} & 42.20 & 73.02 & \underline{86.97} & 42.32 & 63.62 \\
 & Per-factor Muon & \underline{54.92} & 78.18 & 70.88 & \textbf{43.60} & \underline{73.70} & 85.75 & 47.87 & 64.99 \\
 & sMuon & 54.68 & 78.07 & 70.96 & \underline{42.80} & 73.53 & 85.84 & 46.84 & 64.67 \\
 & Riemannion & 54.23 & 78.35 & 71.51 & 42.40 & 73.47 & 85.17 & 47.95 & 64.73 \\
 & Manifold-LoRA & 54.66 & \underline{79.11} & \textbf{72.22} & 42.20 & 73.61 & 85.41 & \textbf{48.81} & \underline{65.14} \\
\rowcolor{prolandrow}
 & \textbf{\method{} (ours)} & \textbf{56.20} & \textbf{79.16} & 71.59 & \underline{42.80} & \textbf{73.99} & \textbf{87.37} & \underline{48.72} & \textbf{65.69} \\
\midrule
\multirow[c]{7}{*}{8} & AdamW & 52.29 & 78.84 & 70.09 & \underline{43.00} & 72.99 & 85.14 & \underline{48.98} & 64.47 \\
 & LoRA-Muon & 53.46 & 76.39 & \underline{72.69} & 41.80 & 73.89 & \textbf{87.77} & 43.86 & 64.26 \\
 & Per-factor Muon & 51.87 & 77.97 & 71.82 & 42.60 & 74.34 & 85.57 & \textbf{49.23} & 64.77 \\
 & sMuon & 54.02 & \textbf{79.33} & 72.38 & 41.80 & 74.44 & \underline{86.73} & 48.21 & 65.27 \\
 & Riemannion & \underline{55.62} & \textbf{79.33} & 71.43 & \underline{43.00} & \underline{74.95} & 85.50 & \underline{48.98} & \underline{65.54} \\
 & Manifold-LoRA & 53.71 & 79.16 & 70.88 & 41.80 & 73.66 & 85.50 & 46.50 & 64.46 \\
\rowcolor{prolandrow}
 & \textbf{\method{} (ours)} & \textbf{56.69} & \underline{79.27} & \textbf{72.93} & \textbf{43.60} & \textbf{75.08} & 85.90 & 48.89 & \textbf{66.05} \\
\midrule
\multirow[c]{7}{*}{16} & AdamW & 51.57 & 78.51 & 71.11 & \textbf{43.20} & 72.14 & \textbf{87.00} & 46.50 & 64.29 \\
 & LoRA-Muon & \underline{55.68} & 78.94 & 72.14 & 42.00 & 73.88 & 85.57 & 48.72 & 65.27 \\
 & Per-factor Muon & 54.12 & 78.35 & 71.27 & 41.80 & 74.02 & 86.64 & 49.83 & 65.15 \\
 & sMuon & 55.01 & \textbf{79.49} & \underline{73.48} & 42.60 & 74.55 & 85.78 & 49.06 & \underline{65.71} \\
 & Riemannion & 54.27 & 78.45 & 71.90 & 40.20 & \underline{74.67} & \underline{86.94} & \textbf{50.85} & 65.33 \\
 & Manifold-LoRA & 52.24 & 77.69 & 69.46 & 39.20 & 73.70 & 85.81 & 46.84 & 63.56 \\
\rowcolor{prolandrow}
 & \textbf{\method{} (ours)} & \textbf{56.75} & \underline{79.27} & \textbf{73.80} & \underline{43.00} & \textbf{75.35} & 85.93 & \underline{49.91} & \textbf{66.29} \\
\bottomrule
\end{tabularx}
\endgroup
\end{table}
\begin{table}[H]
\centering
\caption{\textbf{Moonlight-16B-A3B: language benchmarks.} Zero-shot accuracy (\%) at LoRA rank 16. Avg. is the unweighted mean; bold and underline mark the best and second-best scores.}
\label{tab:moonlight-language}
\begingroup
\renewcommand{\baselinestretch}{1}\fontsize{9}{10.5}\selectfont
\setlength{\tabcolsep}{5.5pt}
\renewcommand{\arraystretch}{1.12}
\begin{tabularx}{\linewidth}{>{\raggedright\arraybackslash}Xrrrrrrrr}
\toprule
Method & MMLU & PIQA & Wino & OBQA & Hella & BoolQ & ARC-C & Avg. \\
\midrule
AdamW & 55.36 & 81.07 & 69.77 & 44.60 & 75.00 & 88.01 & 54.44 & 66.89 \\
LoRA-Muon & 63.81 & \textbf{82.48} & 73.80 & 46.40 & \underline{80.84} & \textbf{89.42} & 61.52 & 71.18 \\
Per-factor Muon & \underline{65.56} & \underline{82.05} & \textbf{74.27} & \underline{47.60} & 80.57 & 87.83 & \underline{63.91} & \underline{71.68} \\
sMuon & 61.18 & 81.23 & 72.61 & 46.80 & 77.26 & 69.79 & 60.41 & 67.04 \\
Riemannion & 59.41 & 81.72 & 73.48 & 46.00 & 76.02 & 87.34 & 57.08 & 68.72 \\
Manifold-LoRA & 58.88 & 81.23 & 71.11 & 44.60 & 76.82 & \underline{88.75} & 57.08 & 68.35 \\
\rowcolor{prolandrow}
\textbf{\method{} (ours)} & \textbf{67.85} & 81.99 & \underline{73.88} & \textbf{47.80} & \textbf{81.56} & 87.37 & \textbf{64.51} & \textbf{72.14} \\
\bottomrule
\end{tabularx}
\endgroup
\end{table}

Moonlight was pretrained with Muon, unlike the two AdamW-pretrained 3B backbones. \citet{liu2025muonscalable} report benefits from retaining Muon across pretraining and full-parameter SFT, and \citet{anson2026smuon} find supporting evidence for Muon-based LoRA on Moonlight. Here, \method{} exceeds per-factor Muon, the strongest baseline, by 0.46 percentage points in mean score and LoRA-Muon by 0.96 points. The advantage over the AdamW baseline is more pronounced, reaching 5.25 percentage points in the seven-task mean. These baselines use Muon-style adapter updates on the same pretrained backbone; therefore, the improvement cannot be attributed to the spectral direction alone. Appendix~\ref{app:optimizer-compatibility} reviews studies on optimizer compatibility between pretraining and fine-tuning.

\subsection{The Role of Landing}\label{sec:feasibility-ablation}
We study how Landing affects orthogonality during training and how the resulting constraint violations relate to task accuracy. Appendix~\ref{app:diagnostic-settings} gives the experimental settings.

\begin{figure}[H]
\centering
\includegraphics[width=\linewidth]{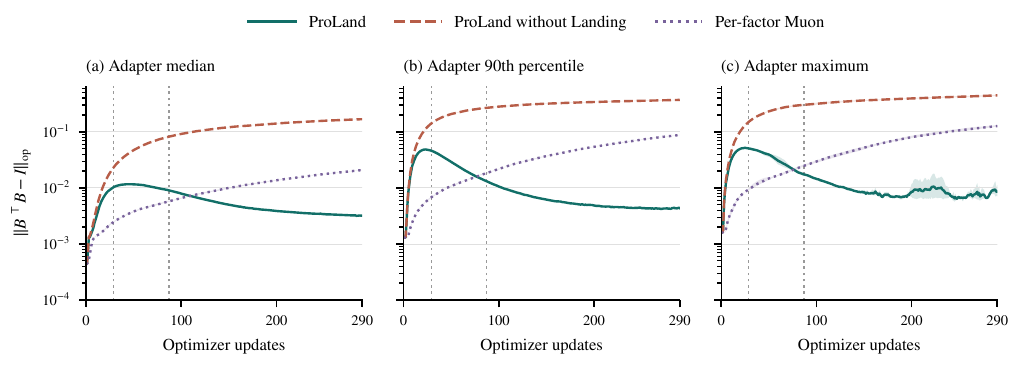}
\caption{\textbf{Landing limits the growth of constraint violation.} Panels show the median, 90th percentile, and maximum of $\|B^\top B-I\|_{\op}$ across adapters. Curves average these statistics across runs; shading shows their range.}
\label{fig:orthogonality}
\end{figure}

\paragraph{Constraint-violation trajectories.}
We compare \method{} with and without Landing, alongside per-factor Muon. In Figure~\ref{fig:orthogonality}, the median constraint violation for \method{} rises initially and then falls. Both controls show increasing median constraint violations late in training. Adding Landing reduces the final median constraint violation, averaged across runs, by 98.1\%. The \method{} trajectory shows gradual recovery of feasibility. Appendix~\ref{app:orthogonality} gives the final constraint violations and their variation across adapters.

\paragraph{Constraint violation and downstream accuracy.}
Table~\ref{tab:obqa-feasibility} compares \method{}, the version without Landing, and the variant with retraction under the same conditions.

Adding Landing reduces the median constraint violation by 92.5\%, the 90th percentile of constraint violation by 96.8\%, and the maximum constraint violation by 96.1\%. It remains effective even in the adapters with the largest violations. Retraction attains smaller constraint violations at the cost of a per-step retraction, while \method{} attains the highest mean accuracy. Both experiments show that Landing keeps the feasibility error small without a retraction at every update.

\begin{table}[H]
\centering
\caption{\textbf{Landing ablation under shared conditions.} Mean $\pm$ sample SD; bold marks the highest accuracy or lowest constraint violation.}
\label{tab:obqa-feasibility}
\begingroup
\renewcommand{\baselinestretch}{1}\fontsize{9}{10.5}\selectfont
\setlength{\tabcolsep}{3pt}
\renewcommand{\arraystretch}{1.16}
\begin{tabularx}{\linewidth}{@{}>{\raggedright\arraybackslash}Xrrrr@{}}
\toprule
& & \multicolumn{3}{c}{Constraint violation ($10^{-3}$) $\downarrow$} \\
\cmidrule(l){3-5}
Method & Acc. (\%) $\uparrow$ & Median & P90 & Maximum \\
\midrule
\rowcolor{prolandrow}
\textbf{\method{}} & $\mathbf{40.73\pm0.69}$ & $7.615\pm0.540$ & $9.389\pm0.732$ & $13.556\pm0.492$ \\
Without Landing & $38.93\pm0.61$ & $101.410\pm0.523$ & $297.497\pm5.759$ & $349.397\pm12.500$ \\
With retraction & $39.67\pm0.76$ & $\mathbf{0.506\pm0.010}$ & $\mathbf{1.282\pm0.025}$ & $\mathbf{1.552\pm0.023}$ \\
\bottomrule
\end{tabularx}
\endgroup
\end{table}
\section{Conclusion}

We introduce \method{}, a retraction-free optimizer for LoRA with $B$ constrained to the Stiefel manifold; it combines product-level spectral updates with coordinated factor updates. The basis-factor direction has no first-order effect on the column orthonormality of $B$, even at infeasible iterates, while a separate Landing correction progressively restores feasibility. An exact low-rank implementation avoids dense product-space optimizer states. Under the stated assumptions and step-size conditions, we establish deterministic and stochastic bounds on product-space stationarity and constraint violation at a common randomized output, assuming exact polar-factor computation. In the main language evaluations, \method{} achieves the highest task-averaged scores among the compared methods in the encoder setting and all seven decoder model--rank configurations. Matched ablations further show that Landing substantially reduces constraint violation.
\label{sec:main-end}
\ifdefined\prolandarxiv
\else
\section*{AI use statement}

We used large language models (LLMs) to assist with language editing, translation, and writing parts of the code (e.g., plotting scripts and Bash scripts). All AI-assisted text and code were reviewed and checked by the authors, who take full responsibility for the final content of this work.
\section*{Reproducibility statement}

The assumptions and complete proofs for all theoretical results are provided in Appendices~A--E. Appendix~B gives the exact low-rank implementation, while Appendices~G--L document the experimental setup, hyperparameters, evaluation protocol, and additional diagnostics needed to reproduce the empirical results.
\fi
\bibliography{references}
\bibliographystyle{iclr2027_conference}
\clearpage
\appendix
\section{Product Geometry and Factor Realizations}\label{app:geometry}

This appendix collects the geometric results underlying the product-space formulation of ProLand. Section~\ref{app:product-geometry} shows that the parameterization with $B$ constrained to the Stiefel manifold preserves the full rank-at-most-$r$ LoRA class and characterizes the first-order product space and its orthogonal projector. Section~\ref{app:stationarity} relates the projected gradient and product-space stationarity measure used in the convergence analysis to the first-order conditions of the factorized constrained problem. Section~\ref{app:factor-realization} shows that every realizable product direction admits a factor realization satisfying the orthogonality condition required by the feasibility analysis, and establishes the reconstruction identity used by the exact low-rank implementation.

\subsection{One-Sided Stiefel Parameterization and Product Geometry}\label{app:product-geometry}

We first prove the expressivity identity stated in Section~\ref{sec:preliminaries}. For any $W\in\R^{m\times n}$ with $1\le k:=\rank(W)\le r$, let $W=U_k\Sigma_kV_k^\top$ be its compact singular value decomposition. Complete $U_k$ to an $m\times r$ matrix with orthonormal columns and set
\[
B=[U_k\ \ U_c]\in\St(m,r),
\qquad
A=
\begin{bmatrix}
\Sigma_kV_k^\top\\
0
\end{bmatrix}.
\]
Then $BA=W$. If $W=0$, take any $B\in\St(m,r)$ and $A=0$. Conversely, every admissible factor pair satisfies $\rank(BA)\le r$. Hence
\[
\{BA:B\in\St(m,r),\ A\in\R^{r\times n}\}
=
\{W\in\R^{m\times n}:\rank(W)\le r\}.
\]
Moreover, for feasible $B$, $(BA)^\top(BA)=A^\top B^\top BA=A^\top A$. Thus $A$ and $BA$ have the same nonzero singular values: the orthogonality constraint normalizes the basis factor $B$ without fixing the scale or singular values of the represented product.

\begin{lemma}[Product differential image and projector]\label{lem:projector}
Let $B\in\R^{m\times r}$ and $A\in\R^{r\times n}$ have arbitrary ranks. Then
\[
\Sspace_{B,A}
=
\range\Diff\Phi(B,A)
=
\{Z\in\R^{m\times n}:(I-P_B)Z(I-P_A)=0\}.
\]
The map
\[
\Pi_{B,A}Z=P_BZ+(I-P_B)ZP_A
\]
is the Frobenius-orthogonal projector onto $\Sspace_{B,A}$. Moreover,
\[
Z\in\Sspace_{B,A}\ \Longrightarrow\ \rank(Z)\le2r,
\qquad
\nrm{\Pi_{B,A}Z}_{\fro}\le\nrm{Z}_{\fro}
\quad\text{for all }Z\in\R^{m\times n}.
\]
\end{lemma}
\begin{proof}
If $Z\in\Sspace_{B,A}$, then $Z=\Delta BA+B\Delta A$ for some factor perturbations. Since $A(I-P_A)=0$ and $(I-P_B)B=0$, $(I-P_B)Z(I-P_A)=0$. Thus
\[
\Sspace_{B,A}\subseteq\{Z:(I-P_B)Z(I-P_A)=0\}.
\]
Conversely, if $(I-P_B)Z(I-P_A)=0$, the four-block decomposition gives
\[
\begin{aligned}
Z
&=P_BZP_A+P_BZ(I-P_A)+(I-P_B)ZP_A\\
&\qquad+(I-P_B)Z(I-P_A)\\
&=P_BZ+(I-P_B)ZP_A\\
&=B(B^\dagger Z)+\bigl((I-P_B)ZA^\dagger\bigr)A.
\end{aligned}
\]
This has the form $B\Delta A+\Delta BA$; hence $Z\in\Sspace_{B,A}$, which proves the characterization.

For arbitrary $Z$, $(I-P_B)(\Pi_{B,A}Z)(I-P_A)=0$. By the characterization above, $\Pi_{B,A}Z\in\Sspace_{B,A}$, hence $\range\Pi_{B,A}\subseteq\Sspace_{B,A}$. For $Y\in\Sspace_{B,A}$,
\[
Y-\Pi_{B,A}Y=(I-P_B)Y(I-P_A)=0.
\]
Therefore $Y=\Pi_{B,A}Y\in\range\Pi_{B,A}$, and hence
\[
\range\Pi_{B,A}=\Sspace_{B,A}.
\]
Since $\Pi_{B,A}$ maps into $\Sspace_{B,A}$ and acts as the identity on this space, $\Pi_{B,A}^2=\Pi_{B,A}$; hence it is a projection onto $\Sspace_{B,A}$.

For arbitrary $Z$, $Z-\Pi_{B,A}Z=(I-P_B)Z(I-P_A)$. For every $Y\in\Sspace_{B,A}$,
\[
\begin{aligned}
\ip{Z-\Pi_{B,A}Z}{Y}
&=\operatorname{tr}\!\left(Z^\top(I-P_B)Y(I-P_A)\right)\\
&=0,
\end{aligned}
\]
where the last equality uses $(I-P_B)Y(I-P_A)=0$. Hence $Z-\Pi_{B,A}Z\in\Sspace_{B,A}^{\perp}$, proving that $\Pi_{B,A}$ is the Frobenius-orthogonal projector onto $\Sspace_{B,A}$.

For $Z\in\Sspace_{B,A}$,
\[
\rank(Z)\le\rank(P_BZ)+\rank((I-P_B)ZP_A)\le2r.
\]
Finally, $\nrm{Z}_{\fro}^2=\nrm{\Pi_{B,A}Z}_{\fro}^2+\nrm{Z-\Pi_{B,A}Z}_{\fro}^2$, hence $\nrm{\Pi_{B,A}Z}_{\fro}\le\nrm{Z}_{\fro}$.
\end{proof}

\subsection{Product-Space and Factor-Space Stationarity}\label{app:stationarity}

\begin{proposition}[Equivalent first-order conditions]\label{prop:stationarity}
Let $B\in\St(m,r)$ and $A\in\R^{r\times n}$, and suppose $f$ is differentiable at $BA$. For $F:=\nabla f(BA)$ and $R:=\Pi_{B,A}F$,
\[
R=0
\quad\Longleftrightarrow\quad
B^\top F=0\ \text{and}\ FA^\top=0.
\]
Under the induced Frobenius metric,
\[
R=0
\quad\Longleftrightarrow\quad
\nabla_A f(BA)=0\ \text{and}\ \grad_B f(BA)=0.
\]
Consequently, at feasible factors, the stationarity measure $\nrm{R}_*$ used in Section~\ref{sec:theory} vanishes exactly at first-order stationary points of the factor problem with $B$ constrained to the Stiefel manifold.
\end{proposition}
\begin{proof}
By Lemma~\ref{lem:projector}, $R=0$ if and only if $F\perp\Sspace_{B,A}$. The following argument applies to arbitrary-rank factors, without requiring $B$ to be feasible. For arbitrary perturbations,
\[
\begin{aligned}
\ip{F}{\Diff\Phi(B,A)[\Delta B,\Delta A]}
&=\ip{F}{\Delta BA+B\Delta A}\\
&=\ip{FA^\top}{\Delta B}+\ip{B^\top F}{\Delta A}.
\end{aligned}
\]
Since $\Delta B$ and $\Delta A$ can be chosen independently, $F\perp\Sspace_{B,A}$ if and only if $FA^\top=0$ and $B^\top F=0$.

The factor gradients are
\[
\nabla_A f(BA)=B^\top F,
\qquad
\grad_B f(BA)=FA^\top-B\sym(B^\top FA^\top).
\]
If $B^\top F=0$, then $B^\top FA^\top=(B^\top F)A^\top=0$, and hence $\grad_B f(BA)=FA^\top$. This proves the second equivalence. Finally, the nuclear norm vanishes if and only if its argument is zero; hence $\nrm{R}_*=0$ has the same zero set.
\end{proof}

\subsection{Factor Realization and Reconstruction}\label{app:factor-realization}

The feasibility analysis in Section~\ref{sec:theory} requires a factor realization whose $B$-component satisfies $B^\top\Delta B=0$. The following result shows that imposing this condition does not restrict the first-order product directions in $\Sspace_{B,A}$.

\begin{lemma}[Compatible factor realization and reconstruction]\label{lem:compatible-realization}
Let $B\in\R^{m\times r}$ and $A\in\R^{r\times n}$ have arbitrary ranks. Then
\[
\Sspace_{B,A}
=
\left\{\Delta BA+B\Delta A:B^\top\Delta B=0\right\}.
\]
Define
\begin{equation}\label{eq:reconstruction}
\mathcal R_{B,A}(M_A,M_B)
=
(B^\dagger)^\top M_A
+
(I-P_B)M_B(A^\dagger)^\top.
\end{equation}
If $M_A=B^\top Z$ and $M_B=ZA^\top$ are the factor coordinates of a common matrix $Z$, then
\[
\mathcal R_{B,A}(B^\top Z,ZA^\top)=\Pi_{B,A}Z.
\]
\end{lemma}
\begin{proof}
The restricted perturbations yield elements of $\Sspace_{B,A}$ by the definition of $\Diff\Phi$. Conversely, given any factor perturbation $(\Delta B,\Delta A)$, set
\[
\Delta B'=(I-P_B)\Delta B,
\qquad
\Delta A'=\Delta A+B^\dagger\Delta BA.
\]
Since $B^\top(I-P_B)=0$,
\[
\begin{aligned}
B^\top\Delta B'&=0,\\
\Delta B'A+B\Delta A'
&=(I-P_B)\Delta BA+B\Delta A+P_B\Delta BA\\
&=\Delta BA+B\Delta A.
\end{aligned}
\]
Thus every element of $\Sspace_{B,A}$ admits a restricted realization, proving the set equality.

For the reconstruction identity, using $(B^\dagger)^\top B^\top=P_B$ and $A^\top(A^\dagger)^\top=P_A$ gives
\[
\begin{aligned}
\mathcal R_{B,A}(B^\top Z,ZA^\top)
&=P_BZ+(I-P_B)ZP_A\\
&=\Pi_{B,A}Z.
\end{aligned}
\]
\end{proof}

\paragraph{Least-squares characterization and the selected factor directions.}
For the current factors and polar factor $Y_t$ in Algorithm~\ref{alg:proland}, the product-space projection has the equivalent least-squares characterization
\[
(D_{A,t}^{\star},D_{B,t}^{\star})
\in
\underset{D_A,D_B}{\arg\min}
\left\|Y_t-\left(B_tD_A+D_BA_t\right)\right\|_{\fro}^2.
\]
Here $D_A\in\R^{r\times n}$ and $D_B\in\R^{m\times r}$. Since $B_tD_A+D_BA_t$ ranges over $\Sspace_t$, Lemma~\ref{lem:projector} gives the unique optimal product direction $\xi_t=\Pi_tY_t$. Indeed, for every factor pair,
\[
\begin{aligned}
\left\|Y_t-(B_tD_A+D_BA_t)\right\|_{\fro}^2
&=\|Y_t-\xi_t\|_{\fro}^2
+\left\|\xi_t-(B_tD_A+D_BA_t)\right\|_{\fro}^2.
\end{aligned}
\]
Thus the product direction is unique, but the minimizing factor pair need not be unique. Algorithm~\ref{alg:proland} selects
\[
D_{A,t}=B_t^\dagger Y_t,
\qquad
D_{B,t}=(I-P_{B_t})Y_tA_t^\dagger.
\]
Using $P_{B_t}=B_tB_t^\dagger$, $P_{A_t}=A_t^\dagger A_t$, and $B_t^\top(I-P_{B_t})=0$ yields
\[
\begin{aligned}
B_tD_{A,t}+D_{B,t}A_t
&=P_{B_t}Y_t+(I-P_{B_t})Y_tP_{A_t}
=\Pi_tY_t=\xi_t,\\
B_t^\top D_{B,t}&=0.
\end{aligned}
\]
These identities hold for arbitrary-rank factors, whether or not $B_t$ is feasible. The selected pair therefore also solves the least-squares problem with the additional condition $B_t^\top D_B=0$, with the same optimal value and product direction. This is a characterization of the existing factor directions, not an additional numerical solve in Algorithm~\ref{alg:proland}.
\section{Exact Low-Rank Implementation}\label{app:backend}

With the reconstruction map $\mathcal R_{B,A}$ in~\eqref{eq:reconstruction}, Algorithm~\ref{alg:proland} admits an exact factor-coordinate realization without forming dense product-space states. Algorithm~\ref{alg:factor-proland} stores the EMA state in thin-factor form, reconstructs both $\Hbar_t$ and the polar-factor input $J_t$ from factor coordinates, and evaluates the polar factor through a core whose row and column dimensions are each at most $2r$.

\begin{algorithm}[H]
\caption{Exact factor-coordinate realization of \method{}}
\label{alg:factor-proland}
\begin{algorithmic}[1]
\Require Initial factors $B_0,A_0$; thin EMA factors
$L_{H,-1},R_{H,-1}$ with
$L_{H,-1}R_{H,-1}^\top=\Hbar_{-1}$ and width at most $2r$;
horizon $T$; $\beta\in[0,1)$, $\rho>0$, steps $\gamma_t>0$

\For{$t=0,\ldots,T-1$}

\Statex
\Statex \textbf{Coordinate Recursions}
\State Obtain factor gradients $g_{A,t},g_{B,t}$ at $(B_t,A_t)$.
\State $\widetilde M_{A,t}\gets
(B_t^\top L_{H,t-1})R_{H,t-1}^\top$;
\Statex \hspace{\algorithmicindent}
$\widetilde M_{B,t}\gets
L_{H,t-1}(R_{H,t-1}^\top A_t^\top)$.
\State $M_{F,t}\gets
\beta\widetilde M_{F,t}+(1-\beta)g_{F,t}$,
$\quad F\in\{A,B\}$;
\Statex \hspace{\algorithmicindent}
$\widehat M_{F,t}\gets
\beta M_{F,t}+(1-\beta)g_{F,t}$,
$\quad F\in\{A,B\}$.

\Statex
\Statex \textbf{Thin Reconstructions}
\State Compute exact pseudoinverses $B_t^\dagger,A_t^\dagger$.
\State $L_{H,t}\gets
[(B_t^\dagger)^\top,\,
M_{B,t}-B_t(B_t^\dagger M_{B,t})]$;
\Statex \hspace{\algorithmicindent}
$R_{H,t}\gets[M_{A,t}^\top,\ A_t^\dagger]$.
\State $L_{J,t}\gets
[(B_t^\dagger)^\top,\,
\widehat M_{B,t}-B_t(B_t^\dagger\widehat M_{B,t})]$;
\Statex \hspace{\algorithmicindent}
$R_{J,t}\gets[\widehat M_{A,t}^\top,\ A_t^\dagger]$.

\Statex
\Statex \textbf{Small-Core Polar Factor}
\State Compute thin QR factorizations
$L_{J,t}=Q_LL_c$ and $R_{J,t}=Q_RR_c$,
with orthonormal columns in $Q_L,Q_R$.
\State $K_c\gets L_cR_c^\top$.
\State Compute the compact SVD
$K_c=U_c\Sigma_cV_c^\top$.
\State $U_{Y,t}\gets Q_LU_c$;
$V_{Y,t}\gets Q_RV_c$.
\Statex \hspace{\algorithmicindent}If $K_c=0$, use zero-rank factors so that $U_{Y,t}V_{Y,t}^\top=0$.

\Statex
\Statex \textbf{Factor Directions and Landing}
\State $D_{A,t}\gets
(B_t^\dagger U_{Y,t})V_{Y,t}^\top$.
\State $D_{B,t}\gets
[U_{Y,t}-B_t(B_t^\dagger U_{Y,t})]
(V_{Y,t}^\top A_t^\dagger)$.
\State $A_{t+1}\gets A_t-\gamma_tD_{A,t}$;
\Statex \hspace{\algorithmicindent}
$B_{t+1}\gets
B_t-\gamma_tD_{B,t}-\rho B_t(B_t^\top B_t-I_r)$.
\State Carry $L_{H,t},R_{H,t}$ to the next iteration.

\EndFor
\end{algorithmic}
\end{algorithm}

\begin{proposition}[Exact factor-coordinate realization at arbitrary rank]
\label{prop:backend}
Consider Algorithms~\ref{alg:proland} and~\ref{alg:factor-proland} with the same initial factors $B_0,A_0$, horizon $T$, and parameters $\beta\in[0,1)$, $\rho>0$, $\gamma_t>0$. Suppose $L_{H,-1}R_{H,-1}^\top=\Hbar_{-1}$ with width at most $2r$, and $g_{A,t}=B_t^\top G_t$, $g_{B,t}=G_tA_t^\top$ for the same product-space input $G_t$ at every $0\le t<T$.
In exact arithmetic, Algorithm~\ref{alg:factor-proland} generates the same
factor iterates as Algorithm~\ref{alg:proland} and satisfies
\[
L_{H,t}R_{H,t}^\top=\Hbar_t,
\qquad
U_{Y,t}V_{Y,t}^\top=\polar(J_t)
\]
at every iteration. The reduced core has dimensions at most $2r\times2r$,
with possibly unequal row and column dimensions; rank-deficient factors and
the zero-core case are included.
\end{proposition}

\begin{proof}
Assume inductively that $L_{H,t-1}R_{H,t-1}^\top=\Hbar_{t-1}$, and define
\[
Z_t:=\beta\Hbar_{t-1}+(1-\beta)G_t.
\]
The supplied initial thin representation and common initial factors give the induction base. At the common factor pair $(B_t,A_t)$, the coordinate recursions give
\[
M_{A,t}=B_t^\top Z_t,
\qquad
M_{B,t}=Z_tA_t^\top.
\]
By the reconstruction identity in Appendix~\ref{app:factor-realization}, using the map in~\eqref{eq:reconstruction},
\[
\mathcal R_{B_t,A_t}(M_{A,t},M_{B,t})
=\Pi_tZ_t
=\Hbar_t.
\]
The second coordinate recursion gives
\[
\widehat M_{A,t}=B_t^\top[\beta Z_t+(1-\beta)G_t],
\qquad
\widehat M_{B,t}=[\beta Z_t+(1-\beta)G_t]A_t^\top.
\]
Applying the same reconstruction identity,
\[
\begin{aligned}
\mathcal R_{B_t,A_t}(\widehat M_{A,t},\widehat M_{B,t})
&=\Pi_t[\beta Z_t+(1-\beta)G_t]\\
&=\beta\Hbar_t+(1-\beta)\Pi_tG_t\\
&=J_t.
\end{aligned}
\]

The thin-factor definitions yield
\[
\begin{aligned}
L_{H,t}R_{H,t}^\top
&=(B_t^\dagger)^\top M_{A,t}
+(I-P_{B_t})M_{B,t}(A_t^\dagger)^\top\\
&=\mathcal R_{B_t,A_t}(M_{A,t},M_{B,t})\\
&=\Hbar_t.
\end{aligned}
\]
The identical calculation with the hatted coordinates gives $L_{J,t}R_{J,t}^\top=J_t$.

Let $L_{J,t}=Q_LL_c$, $R_{J,t}=Q_RR_c$, and $K_c=L_cR_c^\top$. The thin QR factorizations give
\[
J_t=Q_LK_cQ_R^\top.
\]
The row and column dimensions of $K_c$ are each at most $2r$ and may be unequal. If $K_c=U_c\Sigma_cV_c^\top$ is its compact SVD, then
\[
J_t=(Q_LU_c)\Sigma_c(Q_RV_c)^\top.
\]
Since $Q_L,Q_R$ have orthonormal columns and $U_c,V_c$ are the compact singular-vector factors of the core, this is a compact SVD of $J_t$. Therefore,
\[
U_{Y,t}V_{Y,t}^\top
=Q_LU_cV_c^\top Q_R^\top
=Q_L\polar(K_c)Q_R^\top
=\polar(J_t).
\]
If $K_c=0$, then $J_t=0$ and the zero-rank factors agree with $\polar(0)=0$.

Define $Y_t:=U_{Y,t}V_{Y,t}^\top=\polar(J_t)$. Then
\[
D_{A,t}=(B_t^\dagger U_{Y,t})V_{Y,t}^\top=B_t^\dagger Y_t,
\]
and
\[
\begin{aligned}
D_{B,t}
&=[U_{Y,t}-B_t(B_t^\dagger U_{Y,t})](V_{Y,t}^\top A_t^\dagger)\\
&=(I-P_{B_t})Y_tA_t^\dagger.
\end{aligned}
\]
These are exactly the factor directions in Algorithm~\ref{alg:proland}. Therefore the simultaneous $A$- and $B$-updates coincide, closing the induction.
\end{proof}

The dense matrices $\Hbar_t$, $J_t$, and $Y_t$ are represented only through thin factors. Thin matrix products, pseudoinverses, and QR factorizations require $O((m+n)r^2)$ arithmetic up to $r\times r$ core factorizations, while all reduced core factorizations, including the polar-factor computation, require $O(r^3)$. Hence, excluding gradient evaluation, the exact realization requires $O((m+n)r^2+r^3)$ arithmetic operations per iteration and $O((m+n)r)$ additional storage. Only the thin EMA factors $L_{H,t},R_{H,t}$ are carried across iterations.
\section{Pathwise Feasibility and Safe-Region Analysis}\label{app:feasibility}
This appendix proves Proposition~\ref{thm:feasibility}. We first derive the exact Gram recursion and safe-region invariance, then collect the conditioning consequences used later and establish a product-preserving feasible reparameterization. We conclude by showing why pointwise full row rank of $A_t$ is insufficient for a uniform safe-step bound.

\subsection{Exact Gram Recursion and Gram-Polynomial Contraction}\label{app:gram-recursion}
Suppressing the iteration index, consider the basis-factor update
\[
B^+=B-\gamma D-\rho BC=B(I_r-\rho C)-\gamma D,
\qquad C:=B^\top B-I_r,
\qquad K:=B^\top D,
\]
where $D$ denotes the basis-factor direction $D_B$ realizing the optimization update. Since $B^\top B=I_r+C$ and $C=C^\top$, we obtain
\[
\begin{aligned}
C^+
&:=(B^+)^\top B^+-I_r\\
&=(I_r-\rho C)(I_r+C)(I_r-\rho C)-I_r
-\gamma\big[(I_r-\rho C)K+K^\top(I_r-\rho C)\big]
+\gamma^2D^\top D.
\end{aligned}
\]

The first term in the recursion is generated solely by the Landing correction. Expanding it gives
\[
\begin{aligned}
p_\rho(C)
&:=(I_r-\rho C)(I_r+C)(I_r-\rho C)-I_r\\
&=(1-2\rho)C+(\rho^2-2\rho)C^2+\rho^2C^3.
\end{aligned}
\]
Hence, in the absence of the optimization step $(\gamma=0)$, the constraint violation evolves exactly as $C^+=p_\rho(C)$. The linear coefficient $1-2\rho$ captures the local contraction induced by Landing, while the quadratic and cubic terms account for the off-manifold correction. Lemma~\ref{lem:polynomial} turns this local picture into a uniform contraction bound over the safe region $\|C\|_{\op}\le\epsilon$.

The first-order interaction with the optimization direction reveals why the stronger identity $B^\top D_B=0$ is useful. With $K=B^\top D$ as defined above,
\[
-\gamma\big[(I_r-\rho C)K+K^\top(I_r-\rho C)\big]
=-\gamma(K+K^\top)+\gamma\rho(CK+K^\top C).
\]
Suppose only the skew-symmetry condition $B^\top D+D^\top B=0$, the usual Stiefel tangency relation when $B$ lies on the manifold, is imposed. Then $K^\top=-K$; hence the pure $O(\gamma)$ term vanishes, but the mixed term becomes $\gamma\rho(CK-KC)$. This commutator vanishes when $C=0$, but need not vanish away from the manifold. By contrast, the ProLand factor realization satisfies $B_t^\top D_{B,t}=0$, and hence $K=0$ without requiring $C_t=0$. Therefore, all terms that are first order in $\gamma_t$ disappear exactly, and the Gram recursion reduces to
\[
C_{t+1}=p_\rho(C_t)+\gamma_t^2D_{B,t}^\top D_{B,t}.
\]
Thus, the optimization step along $D_{B,t}$ contributes to the constraint violation only through the second-order term $\gamma_t^2D_{B,t}^\top D_{B,t}$.

\begin{lemma}[Contraction of the Gram polynomial]\label{lem:polynomial}
Let $0<\epsilon<1$ and $0<\rho\le1/2$. For any symmetric $C\in\R^{r\times r}$ satisfying $\|C\|_{\op}\le\epsilon$, the Gram polynomial $p_\rho$ defined above satisfies
\[
\|p_\rho(C)\|_{\op}\le q_C\|C\|_{\op},
\qquad
\|p_\rho(C)\|_{\fro}\le q_C\|C\|_{\fro},
\]
where
\[
q_C:=1-2\rho+(2\rho-\rho^2)\epsilon+\rho^2\epsilon^2
=1-(1-\epsilon)(2\rho+\rho^2\epsilon)\in(0,1).
\]
\end{lemma}
\begin{proof}
Let $g_\rho(c):=1-2\rho+(\rho^2-2\rho)c+\rho^2c^2$, so that $p_\rho(c)=c\,g_\rho(c)$. For every $c\in[-\epsilon,\epsilon]$, the condition $0<\rho\le1/2$ gives
\[
\begin{aligned}
|g_\rho(c)|
&\le 1-2\rho+(2\rho-\rho^2)|c|+\rho^2c^2\\
&\le 1-2\rho+(2\rho-\rho^2)\epsilon+\rho^2\epsilon^2=q_C.
\end{aligned}
\]
The bound is attained at $c=-\epsilon$. Moreover, $q_C=1-(1-\epsilon)(2\rho+\rho^2\epsilon)<1$, while its first expression gives $q_C>0$.

Let $C=U\operatorname{diag}(c_1,\ldots,c_r)U^\top$ be an eigendecomposition. Since each $c_i\in[-\epsilon,\epsilon]$ and $p_\rho$ is a polynomial,
\[
p_\rho(C)=U\operatorname{diag}\!\bigl(p_\rho(c_1),\ldots,p_\rho(c_r)\bigr)U^\top.
\]
Therefore,
\[
\|p_\rho(C)\|_{\op}=\max_i|p_\rho(c_i)|\le q_C\max_i|c_i|=q_C\|C\|_{\op},
\]
and similarly
\[
\|p_\rho(C)\|_{\fro}^2=\sum_{i=1}^r|p_\rho(c_i)|^2
\le q_C^2\sum_{i=1}^r|c_i|^2=q_C^2\|C\|_{\fro}^2.
\]
Taking square roots proves the Frobenius-norm bound.
\end{proof}

\subsection{Safe-Region Invariance and Induced Conditioning}
\begin{proof}[Proof of Proposition~\ref{thm:feasibility}]
Under Assumption~\ref{ass:factor}, $\sigma_r(A_t)\ge a_->0$ for every $0\le t\le T$, and hence $\|A_t^\dagger\|_{\op}\le a_-^{-1}$. Since $Y_t=\polar(J_t)$ satisfies $\|Y_t\|_{\op}\le1$ and $I-P_{B_t}$ is an orthogonal projector, the definition $D_{B,t}=(I-P_{B_t})Y_tA_t^\dagger$ gives
\[
\|D_{B,t}\|_{\op}\le a_-^{-1},
\qquad
\|D_{B,t}^\top D_{B,t}\|_{\op}\le a_-^{-2},
\qquad
\|D_{B,t}^\top D_{B,t}\|_{\fro}\le\sqrt r\,a_-^{-2}.
\]
Whenever $\|C_t\|_{\op}\le\epsilon$, the exact Gram recursion and Lemma~\ref{lem:polynomial} give
\[
\begin{aligned}
\|C_{t+1}\|_{\op}&\le q_C\|C_t\|_{\op}+a_-^{-2}\gamma_t^2,\\
\|C_{t+1}\|_{\fro}&\le q_C\|C_t\|_{\fro}+\sqrt r\,a_-^{-2}\gamma_t^2.
\end{aligned}
\]
Under the step-size condition in Proposition~\ref{thm:feasibility}, $\gamma_t\le a_-\sqrt{\epsilon(1-q_C)}$, which gives $a_-^{-2}\gamma_t^2\le(1-q_C)\epsilon$. Thus $\|C_t\|_{\op}\le\epsilon$ implies $\|C_{t+1}\|_{\op}\le q_C\epsilon+(1-q_C)\epsilon=\epsilon$. Together with $\|C_0\|_{\op}\le\epsilon$, induction yields
\[
\|C_t\|_{\op}\le\epsilon,
\qquad 0\le t\le T.
\]
Since $C_t=B_t^\top B_t-I_r$, it follows that
\[
(1-\epsilon)I_r\preceq B_t^\top B_t\preceq(1+\epsilon)I_r,
\]
and therefore
\[
\sqrt{1-\epsilon}\le\sigma_r(B_t)\le\sigma_1(B_t)\le\sqrt{1+\epsilon},
\qquad
\|B_t^\dagger\|_{\op}\le(1-\epsilon)^{-1/2}.
\]

For a constant step size $\gamma_t\equiv\gamma$, iterating the operator-norm recursion gives
\begin{equation}\label{eq:feasbound-app}
\|C_t\|_{\op}
\le q_C^t\|C_0\|_{\op}
+\frac{\gamma^2(1-q_C^t)}{a_-^2(1-q_C)},
\end{equation}
which is the bound stated in Proposition~\ref{thm:feasibility}. For later use, the same argument gives the Frobenius-norm counterpart
\begin{equation}\label{eq:feasbound-fro}
\|C_t\|_{\fro}
\le q_C^t\|C_0\|_{\fro}
+\frac{\sqrt r\,\gamma^2(1-q_C^t)}{a_-^2(1-q_C)}.
\end{equation}
\end{proof}

\paragraph{Empirical control of the $B$ spectrum.}
We examine how Landing controls the spectrum of $B$ during Qwen2.5-3B fine-tuning on CodeAlpaca. We compare Landing strengths under linear-decay and constant learning-rate schedules, including the same update with Landing removed. Training settings are collected in Appendix~\ref{app:diagnostic-settings}.

Figure~\ref{fig:factor-b-residual} shows that Landing suppresses constraint violation under both schedules. With a constant learning rate, the final adapter-median constraint violation, averaged across runs, falls from $1.29\times10^{-1}$ without Landing to $1.69\times10^{-3}$ at $\rho=0.05$, a reduction of approximately $76\times$. The improvement extends to the worst adapter. Through the singular-value bounds in Proposition~\ref{thm:feasibility}, this smaller constraint violation keeps the singular values of $B$ closer to one.

\begin{figure}[!ht]
\centering
\includegraphics[width=\linewidth]{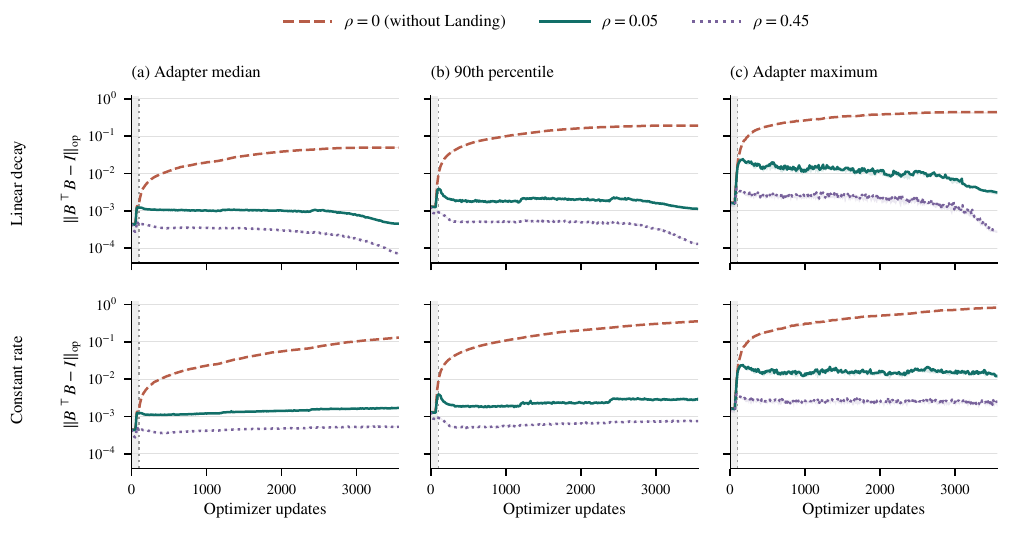}
\caption{\textbf{Landing controls constraint violation of $B$.} Columns show the adapter median, 90th percentile, and maximum; rows compare linear-decay and constant learning-rate schedules. Median and percentile curves show means across runs; maximum curves show the worst run and adapter. Shading spans the run-to-run range, and gray regions mark warmup.}
\label{fig:factor-b-residual}
\end{figure}
\FloatBarrier

\subsection{Additional Conditioning Consequences}
Proposition~\ref{thm:feasibility} already provides uniform conditioning of $B_t$. We record several additional consequences used in the subsequent analysis.

\begin{proposition}[Product conditioning and Stiefel distance]\label{prop:conditioning}
Under the conditions of Proposition~\ref{thm:feasibility}, for every $0\le t\le T$,
\[
\left\|(B_t^\top B_t)^{-1}-I_r\right\|_{\op}
\le\frac{\|C_t\|_{\op}}{1-\|C_t\|_{\op}}.
\]
Moreover,
\begin{equation}\label{eq:distance}
\frac{\|C_t\|_{\fro}}{1+\sqrt{1+\epsilon}}
\le \operatorname{dist}_{\fro}(B_t,\operatorname{St}(m,r))
\le\frac{\|C_t\|_{\fro}}{1+\sqrt{1-\epsilon}}.
\end{equation}
For $i=1,\ldots,r$, the singular values of $W_t=B_tA_t$ satisfy
\[
\sqrt{1-\epsilon}\,\sigma_i(A_t)
\le\sigma_i(W_t)
\le\sqrt{1+\epsilon}\,\sigma_i(A_t).
\]
Consequently, Assumption~\ref{ass:factor} yields
\[
\sigma_r(W_t)\ge\sqrt{1-\epsilon}\,a_-,
\qquad
\|W_t\|_{\op}\le\sqrt{1+\epsilon}\,A_{\max},
\qquad
A_t=B_t^\dagger W_t.
\]
\end{proposition}
\begin{proof}
Let $c_i$ denote an eigenvalue of $C_t=B_t^\top B_t-I_r$. Proposition~\ref{thm:feasibility} gives $|c_i|\le\|C_t\|_{\op}<1$. The corresponding eigenvalue of $(B_t^\top B_t)^{-1}-I_r$ is $-c_i/(1+c_i)$, and hence
\[
\left|\frac{c_i}{1+c_i}\right|\le\frac{\|C_t\|_{\op}}{1-\|C_t\|_{\op}}.
\]

Let $s_i$ denote the singular values of $B_t$. The nearest Stiefel point is the polar factor of $B_t$; therefore $\operatorname{dist}_{\fro}^2(B_t,\operatorname{St}(m,r))=\sum_{i=1}^r(s_i-1)^2$. Since $|s_i-1|=|s_i^2-1|/(s_i+1)$ and Proposition~\ref{thm:feasibility} gives $\sqrt{1-\epsilon}\le s_i\le\sqrt{1+\epsilon}$,
\[
\frac{|s_i^2-1|}{1+\sqrt{1+\epsilon}}
\le |s_i-1|
\le\frac{|s_i^2-1|}{1+\sqrt{1-\epsilon}}.
\]
Squaring and summing over $i$ gives~\eqref{eq:distance}.

Finally, $W_t^\top W_t=A_t^\top B_t^\top B_tA_t$, and therefore
\[
(1-\epsilon)A_t^\top A_t\preceq W_t^\top W_t\preceq(1+\epsilon)A_t^\top A_t.
\]
Eigenvalue monotonicity yields the stated singular-value bounds. Combining them with Assumption~\ref{ass:factor} gives the uniform bounds on $\sigma_r(W_t)$ and $\|W_t\|_{\op}$. Proposition~\ref{thm:feasibility} also ensures that $B_t$ has full column rank; hence $B_t^\dagger W_t=B_t^\dagger B_tA_t=A_t$.
\end{proof}

\subsection{Product-Preserving Feasible Reparameterization}\label{app:feasible-reparameterization}
This subsection establishes a product-preserving feasible reparameterization within the safe region. The result uses only the upper bound $\|A\|_{\op}\le A_{\max}$ and remains valid even if the lower-regularity condition $\sigma_r(A)\ge a_-$ in Assumption~\ref{ass:factor} fails.

\begin{proposition}[Product-preserving feasible reparameterization]\label{prop:certificate}
Let $B\in\R^{m\times r}$ and $A\in\R^{r\times n}$ satisfy $\|B^\top B-I_r\|_{\op}\le\epsilon<1$ and $\|A\|_{\op}\le A_{\max}$, and suppose $f$ is differentiable at $BA$. Write $C:=B^\top B-I_r$ and $R:=\Pi_{B,A}\nabla f(BA)$, and define
\begin{equation}\label{eq:certificate}
\widehat B=B(B^\top B)^{-1/2},
\qquad
\widehat A=(B^\top B)^{1/2}A.
\end{equation}
Then
\[
\widehat B^\top\widehat B=I_r,
\qquad
\widehat B\widehat A=BA,
\qquad
\Pi_{\widehat B,\widehat A}=\Pi_{B,A}.
\]
Defining $\widehat R:=\Pi_{\widehat B,\widehat A}\nabla f(\widehat B\widehat A)$, we further have
\[
\widehat R=R,
\qquad
f(\widehat B\widehat A)=f(BA).
\]
Moreover,
\begin{equation}\label{eq:certdistance}
\|B-\widehat B\|_{\fro}\le\frac{\|C\|_{\fro}}{1+\sqrt{1-\epsilon}},
\qquad
\|A-\widehat A\|_{\fro}\le\frac{A_{\max}\|C\|_{\fro}}{1+\sqrt{1-\epsilon}},
\end{equation}
and, at the feasible pair $(\widehat B,\widehat A)$,
\begin{equation}\label{eq:certgrad}
\|\nabla_{\widehat A}f(\widehat B\widehat A)\|_{\fro}\le\|R\|_*,
\qquad
\|\grad_{\widehat B}f(\widehat B\widehat A)\|_{\fro}
\le\sqrt{1+\epsilon}\,A_{\max}\|R\|_*,
\end{equation}
where $\grad_{\widehat B}$ is defined as in Proposition~\ref{prop:stationarity}.
\end{proposition}
\begin{proof}
Set $G:=B^\top B=I_r+C$. Since $\|C\|_{\op}\le\epsilon<1$, $(1-\epsilon)I_r\preceq G\preceq(1+\epsilon)I_r$; in particular, $G$ is positive definite. Direct multiplication gives
\[
\widehat B^\top\widehat B=G^{-1/2}GG^{-1/2}=I_r,
\qquad
\widehat B\widehat A=BG^{-1/2}G^{1/2}A=BA.
\]
Because $G^{1/2}$ and $G^{-1/2}$ are invertible, $\widehat B$ and $B$ have the same column space, while $\widehat A$ and $A$ have the same row space. Hence $P_{\widehat B}=P_B$ and $P_{\widehat A}=P_A$, and the product-projector formula in Lemma~\ref{lem:projector} gives $\Pi_{\widehat B,\widehat A}=\Pi_{B,A}$. Together with product preservation,
\[
\widehat R=\Pi_{\widehat B,\widehat A}\nabla f(\widehat B\widehat A)
=\Pi_{B,A}\nabla f(BA)=R.
\]

For every eigenvalue $c\in[-\epsilon,\epsilon]$ of $C$,
\[
|\sqrt{1+c}-1|=\frac{|c|}{\sqrt{1+c}+1}
\le\frac{|c|}{1+\sqrt{1-\epsilon}}.
\]
Applying this inequality spectrally gives $\|G^{1/2}-I_r\|_{\fro}\le\|C\|_{\fro}/(1+\sqrt{1-\epsilon})$. Since $B=\widehat BG^{1/2}$ and $\widehat B^\top\widehat B=I_r$, $\|B-\widehat B\|_{\fro}=\|G^{1/2}-I_r\|_{\fro}$. Also, $\widehat A-A=(G^{1/2}-I_r)A$; hence
\[
\|\widehat A-A\|_{\fro}
\le\|G^{1/2}-I_r\|_{\fro}\|A\|_{\op}
\le\frac{A_{\max}\|C\|_{\fro}}{1+\sqrt{1-\epsilon}}.
\]

Finally, let $F:=\nabla f(BA)$. Since $R=\Pi_{B,A}F$, the projector identity gives $F-R=(I-P_B)F(I-P_A)$. Preservation of the factor spaces therefore implies $\widehat B^\top(F-R)=0$ and $(F-R)\widehat A^\top=0$. By the chain rule, $\nabla_{\widehat A}f(\widehat B\widehat A)=\widehat B^\top F=\widehat B^\top R$, and hence $\|\nabla_{\widehat A}f(\widehat B\widehat A)\|_{\fro}\le\|R\|_{\fro}\le\|R\|_*$. The $\widehat B$-component of the Riemannian gradient is the Frobenius-orthogonal projection of $F\widehat A^\top=R\widehat A^\top$ onto $T_{\widehat B}\operatorname{St}(m,r)$. Consequently,
\[
\|\grad_{\widehat B}f(\widehat B\widehat A)\|_{\fro}
\le\|R\widehat A^\top\|_{\fro}
\le\|R\|_{\fro}\|\widehat A\|_{\op}
\le\sqrt{1+\epsilon}\,A_{\max}\|R\|_*,
\]
where $\|\widehat A\|_{\op}=\|G^{1/2}A\|_{\op}\le\sqrt{1+\epsilon}\,A_{\max}$.
\end{proof}

Thus, constraint violation controls the distance to an exactly feasible factorization representing the same product, while the projected gradient $R$ controls first-order stationarity at that feasible representation. This reparameterization is used only in the analysis and is not an additional step of Algorithm~\ref{alg:proland}.

\subsection{Why Pointwise Full Row Rank of $A_t$ Is Insufficient}\label{app:Awindow}
The quantitative lower bound is stronger than pointwise full row rank because both the factor realization and the variation of $P_{A_t}=A_t^\dagger A_t$ depend on $A_t^\dagger$. This mirrors the inverse-smallest-singular-value sensitivity of tangent-space projections in fixed-rank optimization identified by \citet{savostianova2023robust}. The following counterexample shows that requiring each $A_t$ to have full row rank is insufficient for a uniform step-size bound that preserves a fixed safe region.

\begin{proposition}[Counterexample under pointwise full row rank]\label{prop:counterexample}
Fix $0<\epsilon<1$ and any $\gamma>0$. There exist factors with $m=n=2$, $r=1$, $B\in\operatorname{St}(2,1)$, and full-row-rank $A$ such that one ProLand step with exact gradients and $\beta=0$ satisfies
\[
\|(B^+)^\top B^+-I_r\|_{\op}>\epsilon.
\]
Hence, without a positive uniform lower bound on $\sigma_r(A)$, pointwise full row rank alone cannot provide a positive safe step size that is uniform over all admissible $A$.
\end{proposition}
\begin{proof}
Take $B=e_1$, $A=ae_1^\top$ with $a>0$, and $f(W)=\frac12(W_{21}+1)^2$. Then $W=BA=ae_1e_1^\top$ and $F:=\nabla f(W)=e_2e_1^\top$. Since $P_B=P_A=e_1e_1^\top$, the product-projector formula gives $\Pi_{B,A}F=F$. With $\beta=0$, we therefore have $J=F$ and $Y=\polar(J)=F$. Consequently,
\[
D_A=B^\dagger Y=0,
\qquad
D_B=(I-P_B)YA^\dagger=\frac1a e_2,
\qquad
C=B^\top B-I_r=0.
\]
The Landing correction therefore vanishes at this point, and one step gives
\[
B^+=B-\gamma D_B-\rho BC=e_1-\frac{\gamma}{a}e_2,
\qquad
C^+=(B^+)^\top B^+-1=\frac{\gamma^2}{a^2}.
\]
For any fixed $\gamma>0$, choosing $0<a<\gamma/\sqrt{\epsilon}$ gives $C^+>\epsilon$. Nevertheless, $A=ae_1^\top$ has full row rank for every $a>0$, while $\sigma_r(A)=a$ and $\|A^\dagger\|_{\op}=a^{-1}$. Thus the pseudoinverse can become arbitrarily large while full row rank is preserved. Pointwise full row rank therefore does not yield a uniform safe-step bound, which motivates the uniform lower bound $\sigma_r(A_t)\ge a_->0$ in Assumption~\ref{ass:factor}.
\end{proof}

\begin{figure}[!ht]
\centering
\includegraphics[width=\linewidth]{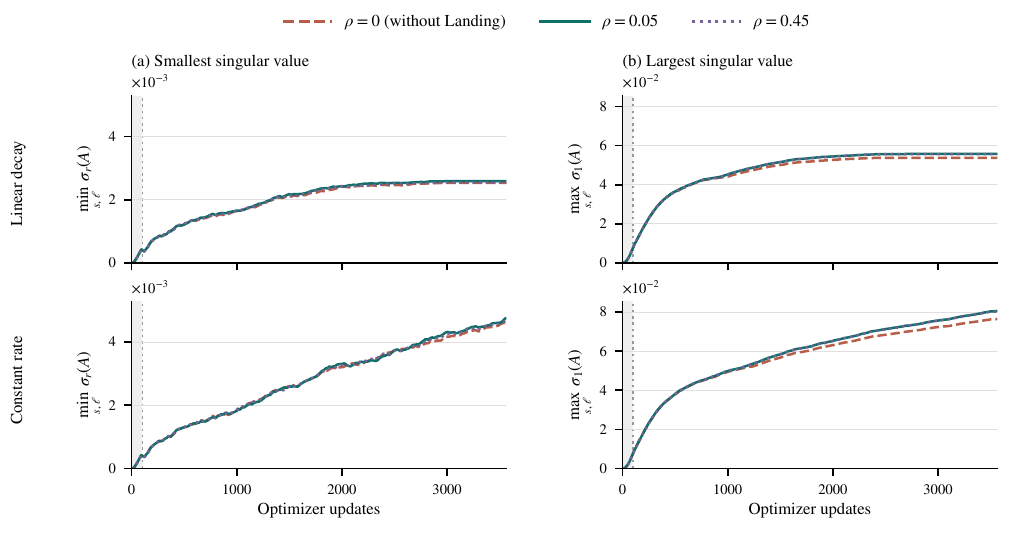}
\caption{\textbf{Extremal singular values of $A$ during fine-tuning.} Curves take extrema over all adapters and runs. The gray region marks warmup; the reported positive lower bound applies to recorded states after warmup. Colors and line styles match Figure~\ref{fig:factor-b-residual}.}
\label{fig:factor-a-spectrum}
\end{figure}

\paragraph{Observed singular-value bounds for $A$.}
Proposition~\ref{prop:counterexample} explains why a quantitative lower bound on $\sigma_r(A)$ enters the safe-step condition. We examine this assumption in the same training runs as in Figure~\ref{fig:factor-b-residual}. Across all recorded post-warmup states, adapters, and runs, we observe $\sigma_r(A)\ge3.65\times10^{-4}$ and $\sigma_1(A)\le8.06\times10^{-2}$, with a maximum condition number below $4.45$. These observations support the $A$-factor regularity assumption over this training window. Figure~\ref{fig:factor-a-spectrum} shows the full trajectories: the extrema stabilize under linear decay and continue to grow under the constant learning rate while remaining within the observed bounds.

\FloatBarrier
\section{Product Descent and Polar-Factor Input Tracking}\label{app:tracking}
This appendix develops the finite-step descent and input-tracking estimates used in Theorem~\ref{thm:convergence}. Section~D.1 derives the exact product increment and one-step descent bound; Section~D.2 controls the variation of the product-space projector and projected gradient; Sections~D.3--D.4 analyze the momentum-state tracking error and its effect on the polar-factor input. Throughout, we write $W_t:=B_tA_t$, $F_t:=\nabla f(W_t)$, $\Delta_t:=J_t-R_t$, and $\bar\gamma:=a_-\sqrt{\epsilon(1-q_C)}$.

\subsection{Exact Product Increment and One-Step Descent}\label{app:product-descent}
Let Algorithm~\ref{alg:proland} run under Assumptions~\ref{ass:objective}--\ref{ass:factor} and the conditions of Proposition~\ref{thm:feasibility}, with a constant optimization step $\gamma_t\equiv\gamma\in(0,\bar\gamma]$.

We first record the exact product increment generated by the simultaneous factor updates. From
\[
A_{t+1}=A_t-\gamma D_{A,t},
\qquad
B_{t+1}=B_t-\gamma D_{B,t}-\rho B_tC_t,
\]
together with~\eqref{eq:stiefel-compatible-realization}, we obtain
\begin{equation}\label{eq:product}
W_{t+1}-W_t=-\gamma\xi_t+Q_t,
\qquad
Q_t=\gamma^2D_{B,t}D_{A,t}-\rho B_tC_t(A_t-\gamma D_{A,t}).
\end{equation}
Proposition~\ref{thm:feasibility} and Assumption~\ref{ass:factor}, together with the rank bounds of the factor directions, give the bounds on $D_{A,t}$ and $D_{B,t}$ below. Lemma~\ref{lem:projector} gives $\operatorname{rank}(J_t)\le2r$; the rank identity for the polar factor and Frobenius nonexpansiveness of $\Pi_t$ then give the bound on $\xi_t$:
\[
\|D_{A,t}\|_{\op}\le(1-\epsilon)^{-1/2},
\qquad
\|D_{A,t}\|_{\fro}\le\sqrt{\frac r{1-\epsilon}},
\qquad
\|D_{B,t}\|_{\fro}\le\frac{\sqrt r}{a_-},
\qquad
\|\xi_t\|_{\fro}\le\sqrt{2r}.
\]
Consequently,
\[
\|Q_t\|_{\fro}
\le\frac{\sqrt r}{a_-\sqrt{1-\epsilon}}\gamma^2
+\rho\sqrt{1+\epsilon}\,\|C_t\|_{\fro}
\left(A_{\max}+\frac{\gamma}{\sqrt{1-\epsilon}}\right).
\]
Using the Frobenius constraint-violation bound~\eqref{eq:feasbound-fro} from Appendix~\ref{app:feasibility}, we obtain
\begin{equation}\label{eq:Q-envelope}
\|Q_t\|_{\fro}\le C_Q\gamma^2+D_Qq_C^t,
\end{equation}
where
\[
\begin{aligned}
C_Q&:=\frac{\sqrt r}{a_-\sqrt{1-\epsilon}}
+\frac{\rho\sqrt{r(1+\epsilon)}}{a_-^2(1-q_C)}
\left(A_{\max}+\frac{\bar\gamma}{\sqrt{1-\epsilon}}\right),\\
D_Q&:=\rho\sqrt{1+\epsilon}\,\|C_0\|_{\fro}
\left(A_{\max}+\frac{\bar\gamma}{\sqrt{1-\epsilon}}\right).
\end{aligned}
\]
Let $S_Q:=\sqrt{2r}+\bar\gamma C_Q$. Then~\eqref{eq:product} and~\eqref{eq:Q-envelope} imply
\begin{equation}\label{eq:product-envelope}
\|W_{t+1}-W_t\|_{\fro}\le S_Q\gamma+D_Qq_C^t.
\end{equation}

By Proposition~\ref{prop:conditioning} and $\operatorname{rank}(W_t)\le r$,
\[
\|W_t\|_{\fro}\le\sqrt r\,\|W_t\|_{\op}
\le\sqrt{r(1+\epsilon)}\,A_{\max}.
\]
Assumption~\ref{ass:objective} therefore gives the uniform gradient bound
\begin{equation}\label{eq:gradient-envelope}
\|F_t\|_{\fro}\le G_{\max}:=\|F_0\|_{\fro}+2L\sqrt{r(1+\epsilon)}\,A_{\max}.
\end{equation}

\begin{lemma}[One-step product descent]\label{lem:descent}
For every $0\le t<T$,
\begin{equation}\label{eq:descent}
f(W_{t+1})
\le f(W_t)-\gamma\|R_t\|_*
+2\sqrt{2r}\,\gamma\|\Delta_t\|_{\fro}
+C_D\gamma^2+D_Dq_C^t,
\end{equation}
where $C_D:=G_{\max}C_Q+LS_Q^2$ and $D_D:=G_{\max}D_Q+LD_Q^2$.
\end{lemma}
\begin{proof}
Starting from the alignment inequality in Section~\ref{sec:theory} and using $\Delta_t\in\Sspace_t$ (hence $\operatorname{rank}(\Delta_t)\le2r$), we obtain
\[
\langle F_t,\xi_t\rangle_{\fro}
\ge\|R_t\|_*-2\|\Delta_t\|_*
\ge\|R_t\|_*-2\sqrt{2r}\,\|\Delta_t\|_{\fro}.
\]
This argument uses only spectral--nuclear duality and does not require any Lipschitz property of the polar-factor map.

Frobenius smoothness along the actual product-update segment gives
\[
f(W_{t+1})
\le f(W_t)+\langle F_t,W_{t+1}-W_t\rangle_{\fro}
+\frac L2\|W_{t+1}-W_t\|_{\fro}^2.
\]
Substituting~\eqref{eq:product} and the alignment bound above yields
\[
f(W_{t+1})
\le f(W_t)-\gamma\|R_t\|_*
+2\sqrt{2r}\,\gamma\|\Delta_t\|_{\fro}
+G_{\max}\|Q_t\|_{\fro}
+\frac L2\|W_{t+1}-W_t\|_{\fro}^2.
\]
Using~\eqref{eq:Q-envelope},~\eqref{eq:product-envelope}, and
\[
(S_Q\gamma+D_Qq_C^t)^2\le2S_Q^2\gamma^2+2D_Q^2q_C^t
\]
gives~\eqref{eq:descent}.
\end{proof}

\subsection{Projector and Projected-Gradient Variation}
We next control the temporal variation of the reachable product space and the projected gradient.

\begin{lemma}[Orthogonal-projector perturbation]\label{lem:subspace}
For matrices $X,Y$ of the same shape, let $P_X=XX^\dagger$ and $P_Y=YY^\dagger$. Then
\[
\|P_X-P_Y\|_{\op}
\le\min\!\left\{1,\,\max\{\|X^\dagger\|_{\op},\|Y^\dagger\|_{\op}\}\|X-Y\|_{\op}\right\}.
\]
No equal-rank assumption on $X$ and $Y$ is required.
\end{lemma}
\begin{proof}
Write $P=P_X$ and $Q=P_Y$. Since
\[
(P-Q)^2=P(I-Q)P+(I-P)Q(I-P),
\]
with the two terms acting on orthogonal subspaces,
\[
\|P-Q\|_{\op}=\max\{\|(I-Q)P\|_{\op},\|(I-P)Q\|_{\op}\}.
\]
Using $(I-Q)Y=0$ and $(I-P)X=0$,
\[
(I-Q)P=(I-Q)(X-Y)X^\dagger,
\qquad
(I-P)Q=(I-P)(Y-X)Y^\dagger.
\]
This gives the second bound. Since $P$ and $Q$ are orthogonal projectors, $-I\preceq P-Q\preceq I$, and hence $\|P-Q\|_{\op}\le1$.
\end{proof}

\begin{lemma}[Projector and projected-gradient variation along \method{}]\label{lem:drift}
Let Algorithm~\ref{alg:proland} run under Assumptions~\ref{ass:objective}--\ref{ass:factor} and the conditions of Proposition~\ref{thm:feasibility}, with $\gamma_t\equiv\gamma\in(0,\bar\gamma]$. Then, for every $0\le t<T$,
\[
\|\Pi_{t+1}-\Pi_t\|_{\fro\to\fro}\le C_\Pi\gamma+D_\Pi q_C^t,
\qquad
\|R_{t+1}-R_t\|_{\fro}\le C_R\gamma+D_Rq_C^t,
\]
where
\[
\begin{aligned}
C_\Pi&:=\frac{2}{a_-\sqrt{1-\epsilon}}
+\frac{\rho\sqrt{1+\epsilon}\,\bar\gamma}{\sqrt{1-\epsilon}\,a_-^2(1-q_C)},
&
D_\Pi&:=\rho\sqrt{\frac{1+\epsilon}{1-\epsilon}}\,\|C_0\|_{\op},\\
C_R&:=LS_Q+G_{\max}C_\Pi,
&
D_R&:=LD_Q+G_{\max}D_\Pi.
\end{aligned}
\]
\end{lemma}
\begin{proof}
For any matrix $Z$, direct subtraction of the product-space projectors gives
\[
(\Pi_{t+1}-\Pi_t)Z
=(P_{B_{t+1}}-P_{B_t})Z(I-P_{A_t})
+(I-P_{B_{t+1}})Z(P_{A_{t+1}}-P_{A_t}).
\]
Lemma~\ref{lem:subspace}, Proposition~\ref{thm:feasibility}, and Assumption~\ref{ass:factor} therefore imply
\[
\|\Pi_{t+1}-\Pi_t\|_{\fro\to\fro}
\le\frac{\|B_{t+1}-B_t\|_{\op}}{\sqrt{1-\epsilon}}
+\frac{\|A_{t+1}-A_t\|_{\op}}{a_-}.
\]
For the $A$-factor, Proposition~\ref{thm:feasibility} gives
\[
\|A_{t+1}-A_t\|_{\op}=\gamma\|D_{A,t}\|_{\op}
\le\frac{\gamma}{\sqrt{1-\epsilon}}.
\]
For the $B$-factor, $B_t^\top D_{B,t}=0$ eliminates the cross terms exactly:
\[
(B_{t+1}-B_t)^\top(B_{t+1}-B_t)
=\gamma^2D_{B,t}^\top D_{B,t}+\rho^2C_t(I_r+C_t)C_t.
\]
Using Proposition~\ref{thm:feasibility} and $\|D_{B,t}\|_{\op}\le a_-^{-1}$,
\[
\|B_{t+1}-B_t\|_{\op}
\le\frac{\gamma}{a_-}+\rho\sqrt{1+\epsilon}\,\|C_t\|_{\op}.
\]
Hence
\[
\|\Pi_{t+1}-\Pi_t\|_{\fro\to\fro}
\le\frac{2\gamma}{a_-\sqrt{1-\epsilon}}
+\rho\sqrt{\frac{1+\epsilon}{1-\epsilon}}\,\|C_t\|_{\op}.
\]
The constant-step constraint-violation bound~\eqref{eq:feasbound} and $\gamma^2\le\bar\gamma\gamma$ yield
\[
\|\Pi_{t+1}-\Pi_t\|_{\fro\to\fro}\le C_\Pi\gamma+D_\Pi q_C^t.
\]

Finally,
\[
R_{t+1}-R_t=\Pi_{t+1}(F_{t+1}-F_t)+(\Pi_{t+1}-\Pi_t)F_t.
\]
Using Frobenius nonexpansiveness of $\Pi_{t+1}$, Assumption~\ref{ass:objective}, and the D.1 bounds~\eqref{eq:product-envelope} and~\eqref{eq:gradient-envelope},
\[
\begin{aligned}
\|R_{t+1}-R_t\|_{\fro}
&\le L\|W_{t+1}-W_t\|_{\fro}
+G_{\max}\|\Pi_{t+1}-\Pi_t\|_{\fro\to\fro}\\
&\le(LS_Q+G_{\max}C_\Pi)\gamma+(LD_Q+G_{\max}D_\Pi)q_C^t
=C_R\gamma+D_Rq_C^t.
\end{aligned}
\]
\end{proof}

\subsection{Momentum-State Tracking and Noise Control}\label{app:tracking-decomposition}
For the stochastic analysis, use the filtration $\{\Fcal_t\}$ in Assumption~\ref{ass:objective}, generated by the algorithmic history up to iteration $t$, before sampling $G_t$, and define $\zeta_t:=G_t-F_t$. Then $B_t,A_t,W_t,\Pi_t,\Hbar_{t-1},F_t,R_t$ are $\Fcal_t$-measurable, while $\E[\zeta_t\mid\Fcal_t]=0$ and $\E[\|\zeta_t\|_{\fro}^2\mid\Fcal_t]\le\sigma^2$. Throughout Sections~D.3--D.4, assume $\E[\|\Hbar_0-R_0\|_{\fro}^2\mid\Fcal_0]\le U_0<\infty$ almost surely, for a deterministic constant $U_0$ independent of $T$. Write $\Ezero[\cdot]:=\E[\cdot\mid\Fcal_0]$, and define the retained-state tracking error $E_t:=\Hbar_t-R_t$. Since $\Pi_tG_t=R_t+\Pi_t\zeta_t$, the polar-factor input mismatch satisfies
\begin{equation}\label{eq:input-mismatch-decomp}
\Delta_t=\beta E_t+(1-\beta)\Pi_t\zeta_t.
\end{equation}

\begin{lemma}[Tracking-error and input-noise decomposition]\label{lem:tracking-noise-decomp}
Define
\[
V_0:=\beta(\Pi_0\Hbar_{-1}-R_0),
\qquad
N_{-1}:=0,
\qquad
N_0:=(1-\beta)\Pi_0\zeta_0,
\]
and, for $t\ge1$ and $t\ge0$, respectively,
\[
V_t=\beta\Pi_t(V_{t-1}+R_{t-1}-R_t),
\qquad
N_t=\beta\Pi_tN_{t-1}+(1-\beta)\Pi_t\zeta_t.
\]
Then, for every $t\ge0$,
\[
E_t=V_t+N_t,
\qquad
\|V_0\|_{\fro}^2\le U_0.
\]
Moreover, for every $k\ge0$,
\[
\Ezero\|N_k\|_{\fro}^2
\le\frac{1-\beta}{1+\beta}(1-\beta^{2k+2})\sigma^2.
\]
Define further $L_t:=\beta^2\Pi_tN_{t-1}+(1-\beta^2)\Pi_t\zeta_t$. Then
\[
\Delta_t=\beta V_t+L_t,
\qquad
\Ezero\|L_t\|_{\fro}\le\sigma\chi_N(\beta),
\]
where
\begin{equation}\label{eq:input-noise-factor}
\chi_N(\beta)^2
:=\beta^4\frac{1-\beta}{1+\beta}+(1-\beta^2)^2
=\frac{1-\beta}{1+\beta}(1+2\beta-2\beta^3).
\end{equation}
\end{lemma}
\begin{proof}
The retained-state update gives, for $t\ge1$,
\[
E_t=\beta\Pi_t(E_{t-1}+R_{t-1}-R_t)+(1-\beta)\Pi_t\zeta_t.
\]
At $t=0$,
\[
E_0=\beta(\Pi_0\Hbar_{-1}-R_0)+(1-\beta)\Pi_0\zeta_0=V_0+N_0.
\]
The definitions of $V_t$ and $N_t$ therefore imply $E_t=V_t+N_t$ by induction. Since $V_0$ and $\Pi_0$ are $\Fcal_0$-measurable and $\E[\zeta_0\mid\Fcal_0]=0$,
\[
\Ezero\|E_0\|_{\fro}^2=\|V_0\|_{\fro}^2+\Ezero\|N_0\|_{\fro}^2.
\]
The initial-state bound thus gives $\|V_0\|_{\fro}^2\le U_0$.

For the noise component, the $\Fcal_t$-measurability of $N_{t-1}$ and $\Pi_t$ gives
\[
\E\!\left[\left\langle\Pi_tN_{t-1},\Pi_t\zeta_t\right\rangle_{\fro}\middle|\Fcal_t\right]=0.
\]
Hence
\[
\E[\|N_t\|_{\fro}^2\mid\Fcal_t]
\le\beta^2\|N_{t-1}\|_{\fro}^2+(1-\beta)^2\sigma^2.
\]
Taking $\Ezero$ and iterating yields
\[
\Ezero\|N_k\|_{\fro}^2
\le(1-\beta)^2\sigma^2\sum_{j=0}^k\beta^{2j}
=\frac{1-\beta}{1+\beta}(1-\beta^{2k+2})\sigma^2.
\]

Finally, using~\eqref{eq:input-mismatch-decomp}, $E_t=V_t+N_t$, and the recurrence for $N_t$,
\[
\Delta_t=\beta V_t+\beta N_t+(1-\beta)\Pi_t\zeta_t
=\beta V_t+L_t.
\]
The same conditional-centering argument gives
\[
\Ezero\|L_t\|_{\fro}^2
\le\beta^4\Ezero\|N_{t-1}\|_{\fro}^2+(1-\beta^2)^2\sigma^2
\le\sigma^2\chi_N(\beta)^2.
\]
Jensen's inequality then gives $\Ezero\|L_t\|_{\fro}\le\sigma\chi_N(\beta)$.
\end{proof}

\subsection{Averaged Tracking of the Polar-Factor Input}
\begin{lemma}[Averaged tracking of the polar-factor input]\label{lem:input-tracking}
Let Algorithm~\ref{alg:proland} run for $T$ iterations under Assumptions~\ref{ass:objective}--\ref{ass:factor}, the initial-state bound in Section~\ref{app:tracking-decomposition}, and the conditions of Proposition~\ref{thm:feasibility}, with fixed $\beta\in[0,1)$ and $\gamma_t\equiv\gamma\in(0,\bar\gamma]$. Let $C_R,D_R$ be the constants from Lemma~\ref{lem:drift}, and define
\[
M_0:=\sqrt{U_0}+\frac{D_R}{1-q_C}.
\]
Then, almost surely,
\begin{equation}\label{eq:averaged-input-tracking}
\frac1T\sum_{t=0}^{T-1}\Ezero\|\Delta_t\|_{\fro}
\le\frac{\beta M_0}{T(1-\beta)}
+\frac{\beta^2C_R\gamma}{1-\beta}
+\sigma\chi_N(\beta),
\end{equation}
where $\chi_N(\beta)$ is defined in Lemma~\ref{lem:tracking-noise-decomp}. The same bound holds for total expectations after replacing the initial-state-dependent quantities by common deterministic upper bounds.
\end{lemma}
\begin{proof}
Lemma~\ref{lem:drift} gives, for $t\ge1$,
\[
\|R_t-R_{t-1}\|_{\fro}\le C_R\gamma+D_Rq_C^{t-1}.
\]
The recurrence for $V_t$ in Lemma~\ref{lem:tracking-noise-decomp} and Frobenius nonexpansiveness of $\Pi_t$ imply
\[
\|V_t\|_{\fro}
\le\beta\|V_{t-1}\|_{\fro}+\beta(C_R\gamma+D_Rq_C^{t-1}).
\]
Iterating gives
\[
\|V_t\|_{\fro}
\le\beta^t\|V_0\|_{\fro}
+\beta\sum_{j=1}^t\beta^{t-j}(C_R\gamma+D_Rq_C^{j-1}).
\]
The finite geometric sums satisfy
\[
\sum_{t=0}^{T-1}\sum_{j=1}^t\beta^{t-j}\le\frac{T-1}{1-\beta},
\qquad
\sum_{t=0}^{T-1}\sum_{j=1}^t\beta^{t-j}q_C^{j-1}
\le\frac1{(1-\beta)(1-q_C)}.
\]
Hence
\[
\frac1T\sum_{t<T}\beta\|V_t\|_{\fro}
\le\frac{\beta(1-\beta^T)\|V_0\|_{\fro}}{T(1-\beta)}
+\frac{\beta^2C_R\gamma}{1-\beta}
+\frac{\beta^2D_R}{T(1-\beta)(1-q_C)}.
\]
Using $\|V_0\|_{\fro}\le\sqrt{U_0}$ from Lemma~\ref{lem:tracking-noise-decomp}, together with $1-\beta^T\le1$ and $\beta^2\le\beta$, yields
\[
\frac1T\sum_{t<T}\beta\Ezero\|V_t\|_{\fro}
\le\frac{\beta M_0}{T(1-\beta)}+\frac{\beta^2C_R\gamma}{1-\beta}.
\]
Finally, Lemma~\ref{lem:tracking-noise-decomp} gives $\Delta_t=\beta V_t+L_t$ and $\Ezero\|L_t\|_{\fro}\le\sigma\chi_N(\beta)$. Therefore,
\[
\begin{aligned}
\frac1T\sum_{t<T}\Ezero\|\Delta_t\|_{\fro}
&\le\frac1T\sum_{t<T}\beta\Ezero\|V_t\|_{\fro}
+\frac1T\sum_{t<T}\Ezero\|L_t\|_{\fro}\\
&\le\frac{\beta M_0}{T(1-\beta)}
+\frac{\beta^2C_R\gamma}{1-\beta}
+\sigma\chi_N(\beta).
\end{aligned}
\]
\end{proof}
\section{Proofs of the Convergence Rates}\label{app:convergence}

This appendix proves the deterministic and stochastic convergence rates in Theorem~\ref{thm:convergence}. We first combine the one-step product descent bound in Lemma~\ref{lem:descent} with the averaged polar-factor input-tracking bound in Lemma~\ref{lem:input-tracking} to control the averaged product-space stationarity measure $\|R_t\|_*$, and use Proposition~\ref{thm:feasibility} together with~\eqref{eq:feasbound-fro} to control the constraint violation. We then derive the guarantees for the same randomized output from these averaged-stationarity and tail-feasibility bounds.

All iteration indices refer to the regular post-initialization segment described in Section~\ref{sec:theory}. The initial state includes the inherited momentum $\Hbar_{-1}$; reindexing does not restart the algorithm. The analysis assumes regularity throughout this segment and does not establish entry into it from zero initialization. The initialization phase and the feasibility cutoff $\tau_T(b)$ are distinct. For unconditional rates, initial-state-dependent quantities in the bounds below are controlled by common deterministic constants independent of $T$.

\subsection{Finite-Horizon Bounds}
Let $D_0:=f(W_0)-f_*+D_D/(1-q_C)$.

\begin{lemma}[Conditional finite-horizon stationarity bound]\label{lem:finite-stationarity}
Let $\{(B_t,A_t)\}_{t=0}^T$ be generated by Algorithm~\ref{alg:proland} under Assumptions~\ref{ass:objective}--\ref{ass:factor}, the initial-state bound in Section~\ref{app:tracking-decomposition}, and the conditions of Proposition~\ref{thm:feasibility}. Use fixed $\beta\in[0,1)$ and $\gamma_t\equiv\gamma\in(0,\bar\gamma]$. With $\Ezero[\cdot]=\E[\cdot\mid\Fcal_0]$, almost surely,
\[
\frac1T\sum_{t=0}^{T-1}\Ezero\|R_t\|_*
\le\frac{D_0}{T\gamma}+C_D\gamma
+2\sqrt{2r}\left[
\frac{\beta M_0}{T(1-\beta)}
+\frac{\beta^2C_R\gamma}{1-\beta}
+\sigma\chi_N(\beta)
\right].
\]
\end{lemma}

\begin{proof}
Lemma~\ref{lem:descent} gives
\[
f(W_{t+1})\le f(W_t)-\gamma\|R_t\|_*
+2\sqrt{2r}\,\gamma\|\Delta_t\|_{\fro}
+C_D\gamma^2+D_Dq_C^t.
\]
Summing over $t=0,\ldots,T-1$ and using $f(W_T)\ge f_*$ and $\sum_{t=0}^{T-1}q_C^t\le(1-q_C)^{-1}$ gives
\[
\gamma\sum_{t=0}^{T-1}\|R_t\|_*
\le D_0+TC_D\gamma^2
+2\sqrt{2r}\,\gamma\sum_{t=0}^{T-1}\|\Delta_t\|_{\fro}.
\]
Dividing by $T\gamma$, taking $\E[\cdot\mid\Fcal_0]$, and invoking Lemma~\ref{lem:input-tracking} yield the claim. Taking total expectations gives the corresponding unconditional estimate.
\end{proof}

\begin{lemma}[Uniform tail-feasibility bound]\label{lem:tail-feasibility}
Suppose the conditions of Proposition~\ref{thm:feasibility} hold for Algorithm~\ref{alg:proland} with
\[
\gamma_t\equiv\gamma_T=c_\gamma T^{-b},
\qquad c_\gamma>0,\quad b>0,
\]
where $c_\gamma$ and $b$ are fixed. Define
\[
\tau_T(b)=\left\lceil\frac{2b\log T}{|\log q_C|}\right\rceil.
\]
For all sufficiently large $T$,
\[
\sup_{\tau_T(b)\le t\le T}\|C_t\|_{\op}
\le\left(\|C_0\|_{\op}+\frac{c_\gamma^2}{a_-^2(1-q_C)}\right)T^{-2b},
\]
\[
\sup_{\tau_T(b)\le t\le T}\|C_t\|_{\fro}
\le\left(\|C_0\|_{\fro}+\frac{\sqrt r\,c_\gamma^2}{a_-^2(1-q_C)}\right)T^{-2b}.
\]
If $C_0=0$, then
\[
\sup_{0\le t\le T}\|C_t\|_{\op}
\le\frac{c_\gamma^2}{a_-^2(1-q_C)}T^{-2b},
\]
\[
\sup_{0\le t\le T}\|C_t\|_{\fro}
\le\frac{\sqrt r\,c_\gamma^2}{a_-^2(1-q_C)}T^{-2b}.
\]
These bounds hold pathwise and do not require the objective-regularity or stochastic-oracle assumptions.
\end{lemma}

\begin{proof}
The operator-norm bound~\eqref{eq:feasbound} and its Frobenius counterpart~\eqref{eq:feasbound-fro} give
\[
\|C_t\|_{\op}\le q_C^t\|C_0\|_{\op}
+\frac{c_\gamma^2T^{-2b}(1-q_C^t)}{a_-^2(1-q_C)},
\]
\[
\|C_t\|_{\fro}\le q_C^t\|C_0\|_{\fro}
+\frac{\sqrt r\,c_\gamma^2T^{-2b}(1-q_C^t)}{a_-^2(1-q_C)}.
\]
For $t\ge\tau_T(b)$,
\[
q_C^t\le q_C^{\tau_T(b)}\le T^{-2b}.
\]
Substitution and taking the supremum prove the tail bounds. If $C_0=0$, the terms involving the initial constraint violation vanish, giving the full-run bounds.
\end{proof}

\subsection{Deterministic and Stochastic Rates}
\begin{proof}[Proof of Theorem~\ref{thm:convergence}(i)]
For exact gradients, $\sigma=0$. Since $\beta\in[0,1)$ is fixed and $\gamma_T=\theta T^{-1/2}\to0$, we have $\gamma_T\le\bar\gamma$ for all sufficiently large $T$; hence Lemma~\ref{lem:finite-stationarity} applies. With no gradient noise, this lemma gives
\[
\frac1T\sum_{t=0}^{T-1}\|R_t\|_*\le\frac{D_0}{T\gamma_T}+C_D\gamma_T+2\sqrt{2r}\left[\frac{\beta M_0}{T(1-\beta)}+\frac{\beta^2C_R\gamma_T}{1-\beta}\right]=O(T^{-1/2}).
\]
Apply Lemma~\ref{lem:tail-feasibility} with $b=1/2$ and $c_\gamma=\theta$:
\[
\sup_{\tau_T(1/2)\le t\le T}\|C_t\|_{\op}
\le\left(\|C_0\|_{\op}+\frac{\theta^2}{a_-^2(1-q_C)}\right)T^{-1},
\]
where
\[
\tau_T(1/2)=\left\lceil\frac{\log T}{|\log q_C|}\right\rceil=O(\log T).
\]
If $C_0=0$, Lemma~\ref{lem:tail-feasibility} gives the same $O(T^{-1})$ bound over the full run $0\le t\le T$. The rates for the same randomized output follow from the argument at the end of this subsection.
\end{proof}

We next consider stochastic gradients.

\noindent\textbf{Choice of exponents.} To explain the parameter choices in Theorem~\ref{thm:convergence}(ii), let
\[
1-\beta_T=c_\beta T^{-a},\qquad
\gamma_T=c_\gamma T^{-b},\qquad
0<a<b<1,
\]
with fixed $c_\beta,c_\gamma>0$.
For $0\le\beta<1$, \eqref{eq:input-noise-factor} gives
\[
\chi_N(\beta)^2
=\frac{1-\beta}{1+\beta}(1+2\beta-2\beta^3)
\le2(1-\beta).
\]
Thus
\[
\chi_N(\beta_T)\le\sqrt{2c_\beta}\,T^{-a/2}.
\]
The finite-horizon term, product/Landing remainder, tracking transient, momentum drift, and stochastic noise in Lemma~\ref{lem:finite-stationarity} scale as
\[
\begin{gathered}
\frac{D_0}{T\gamma_T}=O(T^{b-1}),\qquad
C_D\gamma_T=O(T^{-b}),\qquad
\frac{\beta_TM_0}{T(1-\beta_T)}=O(T^{a-1}),\\
\frac{\beta_T^2C_R\gamma_T}{1-\beta_T}=O(T^{a-b}),\qquad
\sigma\chi_N(\beta_T)=O(T^{-a/2}).
\end{gathered}
\]
Requiring all five terms to be $O(T^{-p})$ gives
\[
1-b\ge p,\qquad b\ge p,\qquad 1-a\ge p,\qquad b-a\ge p,\qquad \frac a2\ge p.
\]
From $a/2\ge p$ we obtain $a\ge2p$, and $b-a\ge p$ gives $b\ge a+p\ge3p$. Meanwhile, $1-b\ge p$ gives $b\le1-p$. Therefore,
\[
3p\le1-p,\qquad p\le\frac14.
\]
When $p=1/4$,
\[
a\ge\frac12,\qquad b\ge a+\frac14\ge\frac34,\qquad b\le\frac34,
\]
hence, uniquely,
\[
(a,b)=\left(\frac12,\frac34\right).
\]
When all five terms are present, the largest decay exponent certified by this bound is $1/4$, uniquely attained at $a=1/2$ and $b=3/4$.

\begin{proof}[Proof of Theorem~\ref{thm:convergence}(ii)]
Set $a=1/2$ and $b=3/4$, that is,
\[
1-\beta_T=c_\beta T^{-1/2},\qquad
\gamma_T=c_\gamma T^{-3/4}.
\]
For all sufficiently large $T$, $\beta_T\in[0,1)$ and $\gamma_T\le\bar\gamma$; hence Lemmas~\ref{lem:finite-stationarity} and~\ref{lem:tail-feasibility} apply.
The five terms have orders
\[
T^{-1/4},\qquad T^{-3/4},\qquad T^{-1/2},\qquad T^{-1/4},\qquad T^{-1/4}.
\]
Taking total expectations in Lemma~\ref{lem:finite-stationarity} gives
\[
\frac1T\sum_{t=0}^{T-1}\E\|R_t\|_*=O(T^{-1/4}).
\]
Lemma~\ref{lem:tail-feasibility} with $b=3/4$ gives, pathwise,
\[
\sup_{\tau_T(3/4)\le t\le T}\|C_t\|_{\op}
\le\left(\|C_0\|_{\op}+\frac{c_\gamma^2}{a_-^2(1-q_C)}\right)T^{-3/2},
\]
where
\[
\tau_T(3/4)=\left\lceil\frac{3\log T}{2|\log q_C|}\right\rceil=O(\log T).
\]
If $C_0=0$, Lemma~\ref{lem:tail-feasibility} gives the same $O(T^{-3/2})$ bound over the full run $0\le t\le T$. The rates for the same randomized output follow from the argument at the end of this subsection.
\end{proof}

\paragraph{Common randomized outputs.}
For either regime, let $\widehat t$ have the distribution specified in Theorem~\ref{thm:convergence}, with $b=1/2$ or $b=3/4$, respectively. Since $\tau_T(b)=O(\log T)$, the sampling interval is nonempty for all sufficiently large $T$, and $T/(T-\tau_T(b))=1+o(1)$. By independence of the output index and nonnegativity,
\[
\E\|R_{\widehat t}\|_*
=\frac{1}{T-\tau_T(b)}\sum_{t=\tau_T(b)}^{T-1}\E\|R_t\|_*
\le\frac{T}{T-\tau_T(b)}\left(\frac1T\sum_{t=0}^{T-1}\E\|R_t\|_*\right).
\]
Thus restriction to this interval preserves the stationarity rate. Moreover,
\[
\E\|C_{\widehat t}\|_{\op}
\le\E\left[\sup_{\tau_T(b)\le t\le T}\|C_t\|_{\op}\right].
\]
Substituting the corresponding averaged-stationarity and tail-feasibility bounds proves both parts of Theorem~\ref{thm:convergence}. In the deterministic case, the expectations on the left are only over $\widehat t$, while those over individual iterates on the right are unnecessary.
\section{Additional Related Work}\label{app:prior}

LoRA learns a weight increment $\Delta W=BA$ while keeping pretrained weights fixed~\citep{hu2021lora}. Representative extensions modify rank allocation through AdaLoRA~\citep{zhang2023adalora}, use different learning rates for the two factors through LoRA+~\citep{hayou2024loraplus}, or separate weight magnitude and direction through DoRA~\citep{liu2024dora}. These developments address different aspects of low-rank adaptation. Expanding the discussion in Section~\ref{sec:related}, we focus on methods that explicitly design the matrix update induced by the factors, impose geometric constraints on the adapter, or change how feasibility is maintained.

The product-space viewpoint concerns the represented matrix $BA$, whose first-order variation is $D_BA+BD_A$. LoRA-Pro selects factor gradients by matching this induced matrix direction to the full fine-tuning gradient in the Frobenius norm~\citep{wang2025lorapro}. LORO derives joint factor updates from Riemannian descent of the represented matrix for low-rank pretraining, combining periodic exact Riemannian updates with cheaper approximate factor updates~\citep{mo2025loro}. Spectral methods extend this viewpoint using Muon-style polar-factor directions~\citep{jordan2024muon}. In particular, sMuon reconstructs a low-rank product-space gradient from factor gradients and derives factor directions by linearizing a relaxed Muon objective and solving the resulting least-squares problem~\citep{anson2026smuon}. LoRA-TSD instead develops tangent-space spectral descent with a factor-induced realization~\citep{andriianov2026loratsd}. ProLand additionally constrains $B$ to the Stiefel manifold and realizes the projected product direction through coordinated factor updates. For the input $J_t$ in Algorithm~\ref{alg:proland}, the selected product direction is $\xi_t=\Pi_t\polar(J_t)$, the Frobenius orthogonal projection of the polar factor onto the current space $\Sspace_t$ of realizable first-order product variations. The coordinated factor directions realize this projected direction, rather than the unprojected polar factor; see Section~\ref{sec:method} and Appendix~\ref{app:factor-realization}.

The constraint may be imposed on the represented matrix or on its factors. Riemannion and LoRA-TSD optimize the represented update on the fixed-rank manifold $\mathcal M_r:=\{W\in\R^{m\times n}:\rank(W)=r\}$~\citep{bogachev2026riemannion,andriianov2026loratsd}. Skewon addresses a different object: it solves a spectral-norm linear minimization problem over $T_X\St$ for a supplied matrix direction, rather than constructing coordinated updates of a LoRA product~\citep{solonko2026skewon}. Among factor-constrained methods, Stiefel-LoRA uses $B\in\St(m,r)$ with $A$ unconstrained~\citep{park2025stiefellora}, and Manifold-LoRA additionally considers an oblique constraint, which normalizes the columns of $B$ without requiring mutual orthogonality~\citep{zhang2026manifoldlora}. Two-sided formulations use three factors: StelLA writes $USV^\top$ and PoLAR writes $X\Theta Y^\top$, with two Stiefel basis factors and an unconstrained $r\times r$ core~\citep{stella2025,lion2025polar}. ProLand retains the two-factor representation $BA$ and constrains only $B$. As shown in Section~\ref{sec:preliminaries}, this formulation represents the full rank-at-most-$r$ class. This representational statement is separate from the convergence assumptions: under the regularity and safe-region conditions of Section~\ref{sec:theory}, the analyzed post-initialization products have rank $r$. The distinction from direct fixed-rank optimization is therefore not simply an admissible-rank comparison, but the use of coordinated factor directions satisfying $B^\top D_B=0$ while realizing a selected first-order product variation.

Feasibility treatment further distinguishes these approaches. Stiefel-LoRA uses a QR retraction on $B$, whereas StelLA uses polar retractions on its two basis factors~\citep{park2025stiefellora,stella2025}. Riemannion uses a truncated-SVD retraction of the represented matrix, and LoRA-TSD uses a factor-induced retraction on $\mathcal M_r$~\citep{bogachev2026riemannion,andriianov2026loratsd}; these are rank constraints on the product, not orthogonality constraints on either LoRA factor. Skewon also uses a retraction to maintain feasibility of its Stiefel variable~\citep{solonko2026skewon}. In contrast, Landing methods allow intermediate infeasible iterates and apply an explicit correction to recover feasibility~\citep{ablin2022landing,ablin2024stiefel}. Manifold-LoRA applies this mechanism to projected-gradient factor updates and establishes convergence for its SGD-type variant~\citep{zhang2026manifoldlora}; PoLAR uses Landing fields on both basis factors in its practical fine-tuning algorithm~\citep{lion2025polar}. ProLand retains a separate Landing correction on $B$, while exploiting the identity $B_t^\top D_{B,t}=0$ even at infeasible iterates. This identity eliminates the optimization-induced term linear in $\gamma_t$ and the mixed optimization--Landing term in the exact Gram recursion in Section~\ref{sec:theory}. The optimization step therefore contributes only a quadratic perturbation to that recurrence. Appendix~\ref{app:gram-recursion} gives the expansion and explains why the usual skew-symmetry condition on $B^\top D_B$ does not by itself eliminate the mixed term away from feasibility.

Momentum handling and low-rank computation are separate from the choice of feasibility mechanism. Riemannion uses projection transport to move tangent-vector momentum between successive tangent spaces~\citep{bogachev2026riemannion}. Skewon's algorithms accept a supplied direction $M_t$, which may be a gradient or momentum, but do not prescribe a recursion for projecting or transporting a stored momentum state between those spaces~\citep{solonko2026skewon}. ProLand explicitly reprojects its previous state before combining it with the current projected gradient; hence both the updated momentum $\Hbar_t$ and the polar-factor input $J_t$ lie in $\Sspace_t$. This gives rank bounds of at most $2r$ for both matrices. Reduced spectral computations already appear in sMuon and Skewon~\citep{anson2026smuon,solonko2026skewon}; ProLand's implementation contribution is the exact low-rank realization of its complete projected-momentum and Landing iteration. Given an initial momentum factorization of width at most $2r$, Appendix~\ref{app:backend} proves equivalence to Algorithm~\ref{alg:proland} in exact arithmetic, using factor gradients from the same sampled loss and a polar-factor core with at most $2r$ rows and columns, without forming dense product-sized gradients or momentum states. Thus, exactness concerns the implementation of the stated projected iteration. Table~\ref{tab:expanded} summarizes the direction, constraint, and feasibility distinctions discussed above.

\begin{table}[!htbp]
\centering
\caption{\textbf{Structural comparison of related low-rank and Stiefel optimization methods.} A checkmark indicates an explicit component of the reviewed method, not a performance advantage.}
\label{tab:expanded}
\begingroup
\small
\setlength{\tabcolsep}{4pt}
\renewcommand{\arraystretch}{1.25}
\begin{tabularx}{\linewidth}{@{}>{\raggedright\arraybackslash}p{0.25\linewidth}>{\raggedright\arraybackslash}p{0.23\linewidth}>{\centering\arraybackslash}p{0.10\linewidth}>{\centering\arraybackslash}p{0.09\linewidth}>{\raggedright\arraybackslash}X@{}}
\toprule
Method & Manifold constraint & Product direction & Muon-type & Feasibility handling \\
\midrule
LoRA-Pro\newline{\footnotesize\citep{wang2025lorapro}} & $\times$ & $\checkmark$ & $\times$ & --- \\
\addlinespace[3pt]
LORO\newline{\footnotesize\citep{mo2025loro}} & $\checkmark\;W\in\mathcal M_r$ & $\checkmark$ & $\times$ & Periodic rank-$r$ retraction \\
\addlinespace[3pt]
sMuon\newline{\footnotesize\citep{anson2026smuon}} & $\times$ & $\checkmark$ & $\checkmark$ & --- \\
\addlinespace[3pt]
LoRA-TSD\newline{\footnotesize\citep{andriianov2026loratsd}} & $\checkmark\;W\in\mathcal M_r$ & $\checkmark$ & $\checkmark$ & Factor-induced rank-$r$ retraction \\
\addlinespace[3pt]
Riemannion\newline{\footnotesize\citep{bogachev2026riemannion}} & $\checkmark\;W\in\mathcal M_r$ & $\checkmark$ & $\checkmark$ & Truncated-SVD retraction \\
\addlinespace[3pt]
Skewon\newline{\footnotesize\citep{solonko2026skewon}} & $\checkmark\;X\in\St$ & $\times$ & $\checkmark$ & Stiefel retraction \\
\midrule
Stiefel-LoRA\newline{\footnotesize\citep{park2025stiefellora}} & $\checkmark\;B\in\St$ & $\times$ & $\times$ & QR retraction on $B$ \\
\addlinespace[3pt]
StelLA\newline{\footnotesize\citep{stella2025}} & $\checkmark\;U,V\in\St$ & $\times$ & $\times$ & Polar retractions on $U,V$ \\
\addlinespace[3pt]
PoLAR\newline{\footnotesize\citep{lion2025polar}} & $\checkmark\;X,Y\in\St$ & $\times$ & $\times$ & Landing on $X,Y$ \\
\addlinespace[3pt]
Manifold-LoRA\newline{\footnotesize\citep{zhang2026manifoldlora}} & $\checkmark\;B\in\St$\newline or $B\in\mathrm{Ob}$ & $\times$ & $\times$ & Landing on $B$ \\
\midrule
\textbf{\method{}}\newline{\footnotesize(this work)} & $\checkmark\;B\in\St$ & $\checkmark$ & $\checkmark$ & Landing on $B$ \\
\bottomrule
\end{tabularx}
\par\smallskip
\begin{minipage}{\linewidth}
\footnotesize
$\checkmark/\times$: present/absent in the reviewed update. ``Product direction'' means explicit selection or matching of the represented matrix update, not merely dependence of the objective on the factors through their product. ``Muon-type'' refers to spectral update design, not a polar parameterization or polar retraction. $\mathcal M_r$ denotes rank-$r$ matrices; $\mathrm{Ob}$ denotes matrices with unit-norm columns. The StelLA entry describes its gradient-based implementation, and the PoLAR entry describes its practical fine-tuning algorithm. A dash means no separate manifold-restoration step; the low-rank parameterization still imposes a rank-at-most-$r$ bound. Landing permits intermediate infeasibility.
\end{minipage}
\endgroup
\end{table}
\FloatBarrier
\clearpage
\section{Supplementary Language Experiments}\label{app:language-details}

\subsection{Rank-8 Loss Trajectories}\label{app:rank8-loss}
\begin{figure}[H]
\centering
\includegraphics[width=\linewidth]{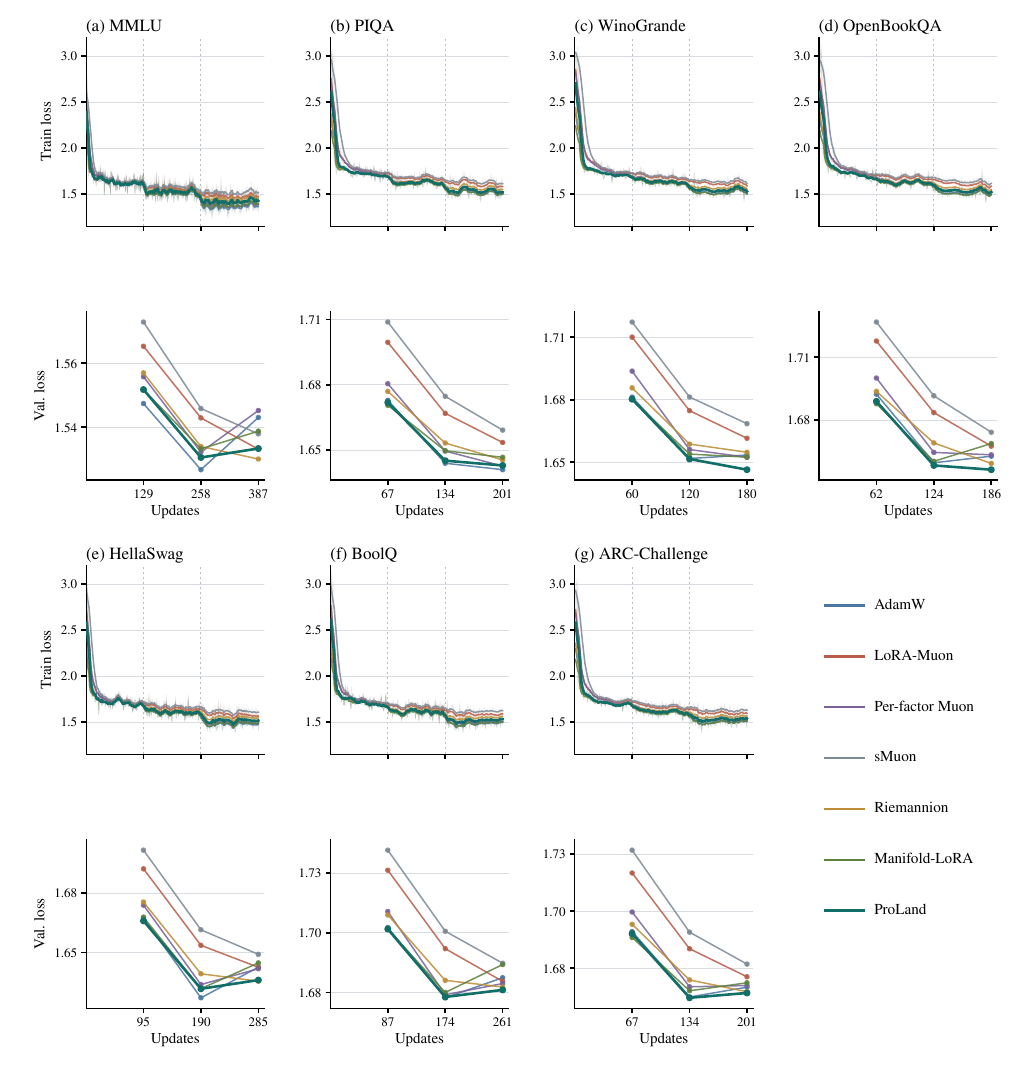}
\caption{\textbf{Qwen2.5-3B rank-8 loss traces.} Each task pairs training loss (upper) with validation loss (lower). Training curves show smoothed and faint raw traces; lines between validation points serve as visual guides.}
\label{fig:qwen-rank8-loss}
\end{figure}

Figure~\ref{fig:qwen-rank8-loss} compares training and validation loss across methods and language tasks on Qwen2.5-3B. The comparison protocol is specified in Appendix~\ref{app:dataset-preparation}.

\subsection{Pretraining and Fine-Tuning Optimizer Compatibility}\label{app:optimizer-compatibility}
\citet[Section~3.5, Table~6]{liu2025muonscalable} cross Muon and AdamW between pretraining and full-parameter SFT using Moonlight checkpoints trained on 1.2T tokens. On their four reported benchmarks, Muon-pretrained checkpoints fine-tuned with Muon outperform the same checkpoints fine-tuned with AdamW.

For low-rank adaptation, \citet[Section~4]{anson2026smuon} explicitly hypothesize that Muon-based LoRA may perform better on Muon-pretrained Moonlight and report partial support across eleven tasks. They also identify Qwen2.5-3B and Llama-3.2-3B as AdamW-pretrained, which is the source of our backbone classification. \citet{qu2026muonfinetune} study optimizer mismatch with controlled pretraining and fine-tuning experiments and find that LoRA can mitigate it by constraining updates.

These findings motivate our choice of models pretrained with different optimizers. The Moonlight comparison in Table~\ref{tab:moonlight-language} uses the same pretrained checkpoint for all adapter methods. On this shared checkpoint, \method{} also has a higher mean score than per-factor Muon and LoRA-Muon, both of which use Muon-style adapter updates.
\section{A Two-Factor Motivation Example}\label{app:motivation}
\paragraph{Two minima of the same rank-one problem.}
Consider fitting $\operatorname{diag}(1.5,1)$ with a rank-one product $W=BA$, assigning weight four to off-diagonal errors:
\begin{equation}
 f(BA)
 =
 \tfrac12\big[
 (w_{11}-1.5)^2
 +(w_{22}-1)^2
 +4w_{12}^2
 +4w_{21}^2
 \big],
 \qquad W=BA.
 \label{eq:motivation-loss}
\end{equation}
To verify its global minimum, write $x=w_{11}$ and $y=w_{22}$. Rank one implies $w_{12}w_{21}=xy$, hence
\[
 f(W)\ge q(x,y):=\tfrac12\big[(x-1.5)^2+(y-1)^2\big]+4|xy|.
\]
Replacing $x,y$ by their absolute values cannot increase $q$. In the positive quadrant its Hessian is $\left(\begin{smallmatrix}1&4\\4&1\end{smallmatrix}\right)$, with a negative eigenvalue; no interior point is a local minimum. Since $q$ is coercive, its minimum lies on an axis. The two axis minima are $(1.5,0)$ and $(0,1)$, with values 0.5 and 1.125. The matrix $W_{\rm good}=\operatorname{diag}(1.5,0)$ attains the lower bound and is globally optimal.
The other matrix, $W_{\rm bad}=\operatorname{diag}(0,1)$, is a strict local minimum under the rank-one constraint. In local coordinates $B=(\delta,\sqrt{1-\delta^2})^\top$ and $A=(a,1+b)$, expansion around $(\delta,a,b)=(0,0,0)$ gives
\[
 f(BA)=1.125+2\delta^2+2a^2-1.5\delta a+\tfrac12b^2+O(\|(\delta,a,b)\|^3).
\]
The quadratic term is positive definite, establishing the claim.
\paragraph{Compared methods.}
We compare Landing against two controls: retraction, and retraction with scale compensation. Initialization and parameter settings are given in Appendix~\ref{app:example-settings}. For the controls, let $\widetilde B=B_t-\gamma_tD_{B,t}$, $\widetilde A=A_t-\gamma_tD_{A,t}$, and $s=\|\widetilde B\|_2$. Retraction sets $(B_{t+1},A_{t+1})=(\widetilde B/s,\widetilde A)$; retraction with scale compensation sets $(B_{t+1},A_{t+1})=(\widetilde B/s,s\widetilde A)$. The latter leaves the product $\widetilde B\widetilde A$ unchanged by retraction.
\paragraph{Observed trajectories and interpretation.}
At the end of the run, \method{} reaches the global minimum with loss 0.5, while both controls reach the local minimum with loss 1.125; all three are feasible and stationary to numerical tolerance. \method{} temporarily leaves the manifold. In panel (a), only the background loss contours minimize over the product scale; panel (b) evaluates the actual iterates.
The construction and settings are collected in Appendix~\ref{app:example-settings}; the example characterizes when the two trajectories separate.
\section{Orthogonality Ablation}\label{app:orthogonality}
This appendix examines variation in constraint violation in the CodeAlpaca Landing ablation of Section~\ref{sec:feasibility-ablation}. The separate OpenBookQA control is reported in Table~\ref{tab:obqa-feasibility}. Training and aggregation settings are collected in Appendix~\ref{app:diagnostic-settings}.

\paragraph{Constraint-violation statistics.}
The largest observed \method{} constraint violation across adapters and updates is 0.053. At the final update, the mean adapter medians are 0.00320, 0.16789, and 0.02079 for \method{}, \method{} without Landing, and per-factor Muon, respectively. \method{} ends lower.

\paragraph{Late-training behavior.}
Late in training, the \method{} median constraint violation continues to decline across runs, while both controls' medians rise.
\section{Factor-Direction Orthogonality in Finite Precision}\label{app:horizontal-check}
The implemented optimization directions satisfy the orthogonality condition $B^\top D_B=0$ to high numerical accuracy at the sampled training states. Sampling and measurement definitions are given in Appendix~\ref{app:diagnostic-settings}.

\begin{figure}[H]
\centering
\includegraphics[width=\linewidth]{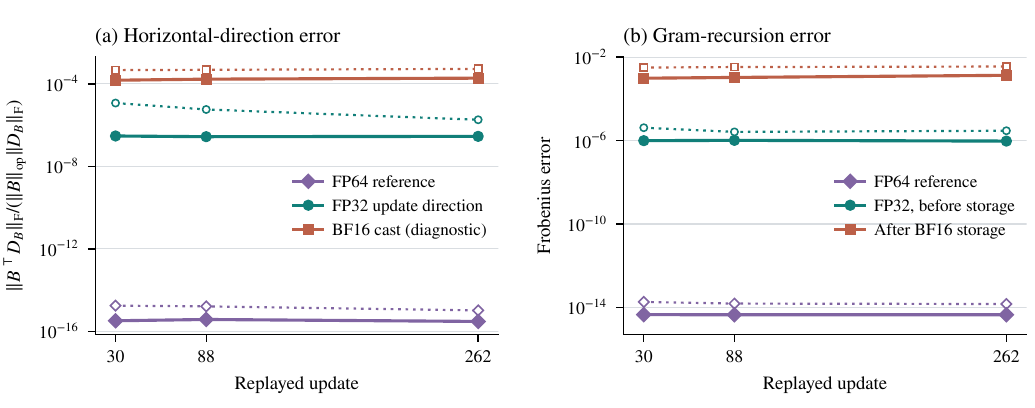}
\caption{\textbf{Factor-direction orthogonality and the effect of parameter storage.} Solid and dotted lines show the median and maximum across runs and adapters. (a) Normalized compatibility error of the FP32 update direction and its BF16 cast. (b) Gram-recursion error before and after BF16 parameter storage.}
\label{fig:horizontal-check}
\end{figure}

\paragraph{Orthogonality of the factor directions.}
Although the sampled bases have constraint violations up to 0.052, the FP32 directions have median compatibility error $e_H=2.88\times10^{-7}$ and maximum $1.19\times10^{-5}$. The implemented directions therefore satisfy the orthogonality condition to high numerical accuracy.
\clearpage
\section{Code-Generation Experiments}\label{app:code}
We also evaluate the same adapter methods on HumanEval, MBPP, and their augmented tests. Training and evaluation settings are collected in Appendix~\ref{app:code-settings}.

\begin{table}[H]
\centering
\caption{\textbf{Code-generation results.} Zero-shot greedy pass@1 (\%). Avg. is the arithmetic mean of the four scores; bold and underline mark the best and second-best scores within each backbone.}
\label{tab:code-generation}
\begingroup
\renewcommand{\baselinestretch}{1}\fontsize{9}{10.5}\selectfont
\setlength{\tabcolsep}{5.5pt}
\renewcommand{\arraystretch}{1.12}
\begin{tabularx}{\linewidth}{>{\raggedright\arraybackslash}Xrrrrr}
\toprule
Method & HumanEval & HumanEval+ & MBPP & MBPP+ & Avg. \\
\midrule
\addlinespace[2pt]
\multicolumn{6}{c}{\textbf{\textit{Qwen2.5-3B}}} \\
\addlinespace[2pt]
AdamW & 54.88 & 51.83 & 66.67 & 57.67 & 57.76 \\
LoRA-Muon & 57.32 & 53.05 & 67.99 & 57.67 & 59.01 \\
Per-factor Muon & 56.71 & 53.05 & \underline{68.25} & 56.88 & 58.72 \\
sMuon & 56.71 & 53.05 & 67.46 & \textbf{58.99} & 59.05 \\
Riemannion & \underline{59.76} & \underline{54.88} & 67.99 & 57.14 & \underline{59.94} \\
Manifold-LoRA & 57.93 & 53.05 & \textbf{69.58} & \textbf{58.99} & 59.89 \\
\rowcolor{prolandrow}
\textbf{\method{} (ours)} & \textbf{60.98} & \textbf{56.71} & 67.20 & \underline{57.94} & \textbf{60.70} \\
\midrule
\addlinespace[2pt]
\multicolumn{6}{c}{\textbf{\textit{Llama-3.2-3B}}} \\
\addlinespace[2pt]
AdamW & 30.49 & 28.05 & \textbf{50.53} & \underline{39.15} & 37.05 \\
LoRA-Muon & 31.71 & 26.22 & 48.41 & 37.83 & 36.04 \\
Per-factor Muon & 31.71 & 25.61 & 49.47 & 38.62 & 36.35 \\
sMuon & \textbf{33.54} & \textbf{29.88} & 49.21 & \underline{39.15} & \textbf{37.94} \\
Riemannion & 31.10 & 27.44 & \underline{50.26} & \textbf{39.68} & 37.12 \\
Manifold-LoRA & \underline{32.93} & \textbf{29.88} & 47.88 & 38.62 & 37.33 \\
\rowcolor{prolandrow}
\textbf{\method{} (ours)} & 31.71 & \underline{28.66} & 49.74 & \textbf{39.68} & \underline{37.45} \\
\bottomrule
\end{tabularx}
\endgroup
\end{table}
\paragraph{Results.}
At rank 8, \method{} achieves the highest average score on Qwen (60.70), outperforming the strongest baseline, Riemannion, by 0.76 percentage points (Table~\ref{tab:code-generation}). On Llama, it ties Riemannion for the highest MBPP+ score (39.68).

\paragraph{Moonlight code results.}
On Moonlight-16B-A3B at rank 16, \method{} has the highest four-metric average (59.89), followed by sMuon (59.37) and per-factor Muon (57.59; Table~\ref{tab:moonlight-code}). LoRA-Muon leads on HumanEval and HumanEval+, while \method{} leads on MBPP and MBPP+.

\begin{table}[H]
\centering
\caption{\textbf{Moonlight-16B-A3B: code benchmarks (\%).} Avg. is the arithmetic mean of the four unrounded pass@1 scores; bold and underline mark the best and second-best scores.}
\label{tab:moonlight-code}
\begingroup
\renewcommand{\baselinestretch}{1}\fontsize{9}{10.5}\selectfont
\setlength{\tabcolsep}{4pt}
\renewcommand{\arraystretch}{1.12}
\begin{tabularx}{\linewidth}{>{\raggedright\arraybackslash}Xrrrrr}
\toprule
Method & HumanEval & HumanEval+ & MBPP & MBPP+ & Avg. \\
\midrule
AdamW & 56.10 & 49.39 & 61.11 & 50.79 & 54.35 \\
LoRA-Muon & \textbf{60.98} & \textbf{52.44} & 62.17 & 52.12 & 56.93 \\
Per-factor Muon & 54.88 & 48.78 & 68.78 & 57.94 & 57.59 \\
sMuon & 57.93 & \underline{51.22} & \underline{69.58} & \underline{58.73} & \underline{59.37} \\
Riemannion & 47.56 & 38.41 & 60.85 & 49.74 & 49.14 \\
Manifold-LoRA & 53.66 & 43.90 & 60.58 & 49.74 & 51.97 \\
\rowcolor{prolandrow}
\textbf{\method{}} & \underline{59.15} & 50.00 & \textbf{71.16} & \textbf{59.26} & \textbf{59.89} \\
\bottomrule
\end{tabularx}
\endgroup
\end{table}
\clearpage
\raggedbottom
\section{Experimental Settings}\label{app:experiments}
\subsection{Shared Settings}\label{app:shared-settings}
We use training seed 2026 for the decoder language and code benchmarks and the loss-trajectory comparisons. The DeBERTa GLUE comparison, CodeAlpaca and OpenBookQA Landing ablations, and factor diagnostics additionally use seeds 1337 and 3407. For the CodeAlpaca orthogonality ablation and factor diagnostics, initialization and data order are paired across settings.

We use the base checkpoints \href{https://huggingface.co/Qwen/Qwen2.5-3B}{Qwen/Qwen2.5-3B}~\citep{qwen2024qwen25}, \href{https://huggingface.co/meta-llama/Llama-3.2-3B}{meta-llama/Llama-3.2-3B}~\citep{meta2024llama32}, and \href{https://huggingface.co/moonshotai/Moonlight-16B-A3B}{moonshotai/Moonlight-16B-A3B}~\citep{liu2025muonscalable}. Table~\ref{tab:training-settings} collects the shared training parameters for the main decoder benchmarks. Qwen and Moonlight use their tokenizers' chat templates; Llama uses a fixed Alpaca-style template.

\begin{table}[H]
\centering
\caption{\textbf{Fine-tuning and decoding settings for Qwen, Llama, and Moonlight.} Experiment-specific choices are listed in the following tables.}
\label{tab:training-settings}
\begingroup
\small
\setlength{\tabcolsep}{4pt}
\renewcommand{\arraystretch}{1.12}
\begin{tabularx}{\linewidth}{@{}l>{\raggedright\arraybackslash}X@{}}
\toprule
Parameter & Value \\
\midrule
Base weights / trainable parameters & Frozen / LoRA factors $A,B$ \\
Adapter targets & All linear modules except the Moonlight MoE routers \\
LoRA scaling / dropout / bias & $\alpha=2r$ / 0 / none \\
Maximum sequence length / packing & 512 / disabled \\
Objective & Response-token-weighted cross-entropy \\
Training precision & BF16; FP32 optimizer arithmetic for \method{} and language-benchmark sMuon \\
Gradient clipping & Global norm 1.0 \\
Benchmark schedule / warmup & Constant / none \\
Benchmark epochs & 3 \\
Adapter ranks & 3B language: 4, 8, 16; 3B code: 8; Moonlight and diagnostics: 16 \\
Evaluation checkpoint & Final adapter of the selected run \\
Code evaluator & EvalPlus 0.3.1; HumanEval+ v0.1.10; MBPP+ v0.2.0 \\
Code decoding & Zero-shot greedy; at most 768 new tokens per problem \\
\bottomrule
\end{tabularx}
\endgroup
\end{table}
\FloatBarrier

\subsection{Data and Evaluation}\label{app:dataset-preparation}
For the decoder benchmarks, we follow sMuon's task and training-data design and its zero-shot evaluation setup~\citep{anson2026smuon}. Each language task has its own adapter. MMLU mixes FLAN and MMLU auxiliary training data in equal proportion (31,129 training and 1,639 holdout rows), and each other language task mixes FLAN, CommonsenseQA, and its task-specific training split in equal proportion (46,694 and 2,458 rows). Source versions and orders are fixed, and smaller sources are cycled deterministically. Code generation and the factor diagnostics use \href{https://github.com/sahil280114/codealpaca}{CodeAlpaca-20k} (19,014 and 1,002 rows). Effective batches range from 240 to 900 sequences per optimizer update across tasks and models. Language evaluation uses frozen zero-shot likelihood prompts in lm-eval, with no evaluation chat template or subsampling. MMLU, WinoGrande, and BoolQ report accuracy; PIQA, OpenBookQA, HellaSwag, and ARC-Challenge report length-normalized accuracy. We use the test splits for MMLU, OpenBookQA, and ARC-Challenge and the validation splits for the other tasks. MMLU aggregates by question count across subjects; the table average weights the seven tasks equally.

\paragraph{Four-task GLUE validation.}\label{app:glue-settings}
We evaluate DeBERTa-v3-base at LoRA rank 16 on the validation splits of RTE, MRPC, STS-B, and CoLA. The task scores are RTE accuracy, the arithmetic mean of MRPC accuracy and F1, the arithmetic mean of STS-B Pearson and Spearman correlations, and CoLA Matthews correlation coefficient (MCC). For each seed, the four-task score is the unweighted mean of these task scores. For each task, Table~\ref{tab:deberta-glue} reports the mean and sample standard deviation across the three seeds listed in Appendix~\ref{app:shared-settings}, using the $n-1$ denominator with $n=3$; the average column reports the mean of the four task scores. This average is specific to these four validation tasks and is not a full GLUE score.

Training uses LoRA on the attention query and value projections only, at rank 16 with scaling $\alpha=32$. The pretrained parameters remain frozen, and the adapters, pooler, and classification head are trained. All methods use the full training set for three epochs with batch size 32 and maximum sequence length 128, a learning rate of $3\times10^{-4}$ for the adapters and the head, and a linear decay after a $6\%$ warmup. Parameters and optimizer states are kept in FP32, and the forward pass uses BF16 autocast. Each method retains its native initialization and update rule; \method{} uses $\rho=0.05$ and weight decay $(0.001,0)$ on $A$ and $B$, while the other methods use weight decay $0.01$; the classification head is trained with AdamW throughout. Every method uses a single preset configuration, no additional hyperparameter search, and the final checkpoint.

\paragraph{Code evaluation.}\label{app:code-settings}
We evaluate HumanEval~\citep{chen2021codex} and MBPP~\citep{austin2021program}, together with their augmented tests in EvalPlus~\citep{liu2023evalplus}. Each problem receives one zero-shot greedy completion, processed with the EvalPlus sanitizer. The four pass@1 scores in each row use the same final adapter and generated answers; their arithmetic mean gives the average score.
\FloatBarrier

\subsection{Parameter Selection and Final Configurations}\label{app:parameter-selection}
\paragraph{3B baselines.}
For each model, rank, and method, we select a language learning rate by accuracy on a deduplicated PIQA training holdout, then use it for all seven task-specific adapters. This selection set contains 794 questions drawn from the training holdout, excluding questions or normalized prompts present in training. For code generation, we select by final response-token-weighted cross-entropy on the CodeAlpaca holdout. Ties prefer the base rate and then the lower rate; code losses within $10^{-12}$ are treated as tied. Table~\ref{tab:selection-grids} gives the search grids and training budgets, and Table~\ref{tab:lr-selection} gives the selected rates. The baseline base rates are taken from sMuon~\citep{anson2026smuon}.

\paragraph{3B \method{} language selection.}
\method{} follows the same protocol as the baselines: for each model and rank, we select a language learning rate from the corresponding three-point grid by accuracy on the deduplicated PIQA training holdout, and use it for all seven task-specific adapters, with the same tie-breaking rule. In the main \method{} decoder language benchmarks, the Landing step is fixed at $\rho=0.05$, and factor $A$ receives no weight decay ($\lambda_A=0$).

\begin{table}[H]
\centering
\caption{\textbf{Candidate grids and selection stages.} All rates are base learning rates before any optimizer-specific scaling.}
\label{tab:selection-grids}
\begingroup
\small
\setlength{\tabcolsep}{4pt}
\renewcommand{\arraystretch}{1.12}
\begin{tabularx}{\linewidth}{@{}l>{\raggedright\arraybackslash}X@{}}
\toprule
Experiment & Candidates / training budget \\
\midrule
3B baselines & $\eta\in\{0.3\eta_0,\eta_0,3\eta_0\}$; 3 epochs per candidate; $\eta_0$ from Table~\ref{tab:lr-selection} \\
\method{} & Same multipliers and budget; language and code $\eta_0=6\times10^{-6}$ \\
Factor-spectrum pilot & $\eta\in\{1.75\times10^{-7},7\times10^{-7},2\times10^{-6},7\times10^{-6}\}$; linear decay, $\rho=0.05$ \\
\bottomrule
\end{tabularx}
\endgroup
\end{table}
\begin{table}[H]
\centering
\caption{\textbf{Final learning rates for the 3B language and code benchmarks and for Moonlight.} 3B language runs use ranks 4, 8, and 16; code runs use rank 8; and Moonlight runs use rank 16.}
\label{tab:lr-selection}
\begingroup
\footnotesize
\setlength{\tabcolsep}{3pt}
\renewcommand{\arraystretch}{1.12}
\begin{tabularx}{\linewidth}{@{}Xrrrrrrrrrrr@{}}
\toprule
& & \multicolumn{3}{c}{Qwen2.5-3B} & \multicolumn{3}{c}{Llama-3.2-3B} & \multicolumn{2}{c}{3B code} & \multicolumn{2}{c}{Moonlight} \\
\cmidrule(lr){3-5}\cmidrule(lr){6-8}\cmidrule(lr){9-10}\cmidrule(lr){11-12}
Method & $\eta_0$ & $r=4$ & $r=8$ & $r=16$ & $r=4$ & $r=8$ & $r=16$ & Qwen & Llama & Lang. & Code \\
\midrule
AdamW & 0.001 & 0.0003 & 0.0003 & 0.0003 & 0.001 & 0.001 & 0.001 & 0.0003 & 0.0003 & 0.001 & 0.001 \\
LoRA-Muon & 1e-5 & 1e-5 & 3e-6 & 3e-5 & 3e-5 & 3e-5 & 1e-5 & 3e-6 & 3e-6 & 6e-5 & 6e-5 \\
Per-factor Muon & 6e-5 & 1.8e-5 & 1.8e-5 & 1.8e-5 & 1.8e-5 & 6e-5 & 6e-5 & 1.8e-5 & 1.8e-5 & 6e-5 & 1.8e-5 \\
sMuon & 7e-6 & 7e-6 & 2.1e-6 & 7e-6 & 2.1e-5 & 2.1e-5 & 2.1e-5 & 2.1e-6 & 7e-6 & 6e-5 & 7e-6 \\
Riemannion & 1e-5 & 1e-5 & 3e-6 & 1e-5 & 3e-5 & 1e-5 & 3e-5 & 3e-6 & 3e-6 & 6e-5 & 6e-5 \\
Manifold-LoRA & 0.001 & 0.0003 & 0.0003 & 0.001 & 0.001 & 0.001 & 0.001 & 0.0003 & 0.0003 & 0.001 & 0.001 \\
\method{} & 6e-6 & 6e-6 & 1.8e-6 & 1.8e-6 & 1.8e-5 & 1.8e-5 & 1.8e-5 & 1.8e-6 & 6e-6 & 6e-5 & 7e-6 \\
\bottomrule
\end{tabularx}
\endgroup
\end{table}
\FloatBarrier

\paragraph{Code and Moonlight.}\label{app:moonlight}
The 3B \method{} code results use the CodeAlpaca-holdout selection. Moonlight uses the method-specific rates from sMuon~\citep{anson2026smuon}; these runs do not use the 3B search grids. Table~\ref{tab:lr-selection} records these rates. Table~\ref{tab:optimizer-settings} gives the fixed 3B optimizer parameters; Moonlight \method{} uses the same momentum and code settings, including $\rho=0.05$ and zero factor weight decay. Manifold-LoRA uses its fixed-step implementation in both Moonlight comparisons.

\begin{table}[H]
\centering
\caption{\textbf{Fixed optimizer settings for the 3B benchmarks.} Baseline weight decay is 0.01 with each implementation's decay rule. AdamW and Manifold-LoRA use Adam coefficients $(0.9,0.999)$ and $\epsilon=10^{-8}$; the other methods use momentum 0.9. These internal settings are held fixed during the learning-rate searches.}
\label{tab:optimizer-settings}
\begingroup
\small
\setlength{\tabcolsep}{4pt}
\renewcommand{\arraystretch}{1.12}
\begin{tabularx}{\linewidth}{@{}l>{\raggedright\arraybackslash}X@{}}
\toprule
Method & Numerical update settings \\
\midrule
AdamW & Standard AdamW update \\
LoRA-Muon & 8 FP32 polar iterations; 7 inverse-root iterations; $\epsilon=10^{-5}$ \\
Per-factor Muon & Independent factor EMAs; 5 BF16 polar iterations \\
sMuon & 12 polar iterations; 8 inverse-root coefficient iterations; $\epsilon=10^{-4}$ \\
Riemannion & Low-rank QR/SVD update \\
Manifold-LoRA & Fixed-step Stiefel--AdamW; restoration coefficient $1/3$ \\
\method{} & Projected Nesterov momentum; numerical iterations and damping as in sMuon; $\rho=0.05$, zero weight decay on both factors \\
\bottomrule
\end{tabularx}
\endgroup
\end{table}

\paragraph{Loss-trajectory comparison.}
Figure~\ref{fig:qwen-rank8-loss} uses fixed three-epoch Qwen runs at rank 8: the baselines use Table~\ref{tab:lr-selection}, and \method{} uses $\rho=0.05$ with zero factor weight decay. We plot nine-update moving averages of training loss and evaluate validation loss after each epoch.
\FloatBarrier

\subsection{Feasibility Ablations and Factor Diagnostics}\label{app:diagnostic-settings}
The constraint-violation trajectory experiment and factor diagnostics fine-tune Qwen2.5-3B on the CodeAlpaca split described above. Table~\ref{tab:diagnostic-settings} collects their training parameters and sampling positions. The separate OpenBookQA control is specified below.

\begin{table}[H]
\centering
\caption{\textbf{Orthogonality and factor-diagnostic settings.} Both use $\beta=0.9$, $\lambda_A=0$, $\lambda_B=0$, orthogonal $B$ and zero $A$ initialization, FP32 updates, and BF16 factors and optimizer states.}
\label{tab:diagnostic-settings}
\begingroup
\small
\setlength{\tabcolsep}{4pt}
\renewcommand{\arraystretch}{1.12}
\begin{tabularx}{\linewidth}{@{}l>{\raggedright\arraybackslash}X>{\raggedright\arraybackslash}X@{}}
\toprule
Parameter & Landing ablation & Factor spectra \\
\midrule
Learning rate & 7e-6 & 7e-7 (peak) \\
Effective batch / epochs & 640 / 10 & 16 / 3 \\
Warmup & None & 100 updates \\
Schedule & Constant & Linear decay to zero; constant \\
Landing strength $\rho$ & 0, 0.05 & 0, 0.05, 0.45 \\
Validation & Each epoch & Four times per epoch \\
Late-training interval & Final 30\% of updates & Final 30\% of updates \\
Direction replay & First update after epochs 1, 3, 9 & Not used \\
Factor measurements & Across all adapters in each run & FP64; 252 adapters per run, 18 runs \\
\bottomrule
\end{tabularx}
\endgroup
\end{table}

\paragraph{CodeAlpaca Landing ablation.}
Figure~\ref{fig:orthogonality} compares \method{}, the same optimizer without Landing, and per-factor Muon. The Landing-free ablation sets $\rho=0$ while leaving the rest of the implementation unchanged. These runs use rank 16, LoRA scaling $\alpha=32$, sequence length 512, and NVIDIA H100 80GB GPUs. A microbatch of 32 sequences with 20 gradient-accumulation steps gives an effective batch of 640; each run completes 290 optimizer updates. Per-factor Muon shares the initial factors, learning rate, and effective weight decay, and uses independent factor EMAs and BF16 polar iterations as in Table~\ref{tab:optimizer-settings}, without Landing or retraction. For each run, we compute the median, 90th percentile, and maximum constraint violation across adapters; the plotted curves average these statistics, and the shading spans their range.

\paragraph{OpenBookQA Landing ablation.}\label{app:obqa-control-settings}
Table~\ref{tab:obqa-feasibility} uses Qwen2.5-3B base with the official OpenBookQA training set, split into 4,709 training and 248 held-out examples (split seed 20260904). Question stems and full correct answers are rendered with the tokenizer's chat template; only response tokens contribute to the loss, with no sequence packing. The frozen model has rank-16 LoRA ($\alpha=32$, dropout 0) on the query, key, value, output, gate, up, and down projections of all 36 layers (252 factor pairs). Within each shared seed, all variants have byte-identical initial factors: $B$ is obtained by QR of an FP32 Gaussian matrix, $A=0$, and momentum is zero.

All variants use three epochs (108 updates), maximum length 256, and batch 128 (microbatch 32, accumulation 4), discarding incomplete accumulation windows. The base learning rate is constant at $7\times10^{-6}$ without warmup; the shared \texttt{match\_rms\_adamw} scaling gives $\eta=7\times10^{-6}\cdot0.2\sqrt{d_{\mathrm{out}}d_{\mathrm{in}}/r}$. Nesterov momentum is 0.9, $(\lambda_A,\lambda_B)=(0.001,0)$, global gradient clipping is 1.0, and Gram inverse-root regularization is $10^{-4}$. Parameters and persistent momentum are BF16, update kernels are FP32, and TF32 is disabled. Each method uses one fixed configuration, zero additional search, and the final checkpoint. Primary results comprise nine H100 runs.

\method{} uses $\rho=0.05$, subtracting $\rho B(B^\top B-I)$ independently of $\eta$. The without-Landing variant sets only $\rho=0$. The with-retraction variant also sets $\rho=0$ and maps the tentative update $\widetilde B$ to $B_{\mathrm{new}}=\widetilde B(\widetilde B^\top\widetilde B)^{-1/2}$ at every step, using FP32 symmetric eigendecomposition. This is the standard polar retraction of the basis factor in formulations with $B$ constrained to the Stiefel manifold. All variants then apply the same projected product-momentum projection using the updated FP32 factors before BF16 storage.

We evaluate all 500 test questions by zero-shot continuation likelihood: the prompt is the question stem, and the candidates are the full option texts, with no evaluation chat template. Table~\ref{tab:obqa-feasibility} reports ordinary accuracy from summed log-likelihoods, computed with FP32 log-softmax on BF16 model outputs. Final constraint violations $\|B^\top B-I\|_{\op}$ are measured in FP64 after BF16 storage. We first compute the median, 90th percentile, and maximum across the 252 factors within each run, then report their means and sample standard deviations across seeds, in units of $10^{-3}$. Accuracy uses the same across-seed summary.

\paragraph{Factor spectra.}
Figures~\ref{fig:factor-b-residual} and~\ref{fig:factor-a-spectrum} compare Landing strengths under linear-decay and constant learning-rate schedules. We choose the peak learning rate in a separate pilot by the mean held-out loss over the final three validation evaluations, with ties resolved in favor of the smaller rate, and keep it fixed across Landing strengths and schedules. For the constraint violation of $B$, median and percentile curves average the per-run statistics, whereas maximum curves take the worst run and adapter; shading shows the range across runs. The $A$ curves take extrema over all runs and adapters. Gray regions denote warmup.

\paragraph{Direction diagnostics.}
Figure~\ref{fig:horizontal-check} replays the first update after selected completed epochs of the Landing-ablation runs, as specified in Table~\ref{tab:diagnostic-settings}. For the optimization direction, excluding Landing, we measure $e_H=\nrm{B^\top D_B}_{\fro}/(\nrm{B}_{\op}\nrm{D_B}_{\fro})$. An FP64 reference applies~\eqref{eq:reconstruction} and~\eqref{eq:stiefel-compatible-realization} to the same stored factors and polar-map inputs. For the exact Gram recurrence in Section~\ref{sec:theory}, we measure $\nrm{C^+-P(C,D_B)}_{\fro}$ before and after BF16 storage, using each adapter's actual step size, and aggregate over runs and adapters.
\FloatBarrier

\subsection{Two-Factor Numerical Example}\label{app:example-settings}
For the example in Appendix~\ref{app:motivation}, all methods use the momentum-projection, polar-map-input, polar-factor, and factor-direction steps of Algorithm~\ref{alg:proland}, recomputed at their own current factors, with $G_t=\nabla f(B_tA_t)$, $\beta=0.9$, and $\Hbar_{-1}=0$. We initialize $B_0=(1/2,\sqrt3/2)^\top$ and $A_0=1.5B_0^\top$. Both factors are updated simultaneously, and each method projects its previous EMA state into its current differential image. The shared schedule is
\[
 \gamma_t=0.2\,(0.97)^{\max\{0,t-100\}},\qquad t=0,\ldots,799.
\]
\method{} uses $\rho=0.3$. This decaying schedule is chosen for the illustration and differs from the horizon-tuned schedule of the convergence bounds.

\end{document}